\pdfoutput=1
\documentclass[conference]{IEEEtran}
\IEEEoverridecommandlockouts
\usepackage{amsmath,amssymb,amsfonts}
\usepackage{balance}
\usepackage{algorithmic}
\usepackage{graphicx}
\usepackage{textcomp}
\usepackage[table]{xcolor}
\usepackage{stfloats}
\usepackage{xcolor}
\usepackage{threeparttable}
\usepackage{color}
\usepackage{graphicx}
\usepackage{balance}
\usepackage[caption=false,font=normalsize,labelfont=sf,textfont=sf]{subfig}
\usepackage[normalem]{ulem}
\useunder{\uline}{\ul}{}
\usepackage{multirow}
\def\BibTeX{{\rm B\kern-.05em{\sc i\kern-.025em b}\kern-.08em
    T\kern-.1667em\lower.7ex\hbox{E}\kern-.125emX}}
\begin{document}

\title{Towards Lightweight Time Series Forecasting: a Patch-wise Transformer with Weak Data Enriching\\
    \thanks{$^*$ Corresponding author.\\$^{\dagger}$ Affiliated with The Shaanxi Key Laboratory of Clothing Intelligence, China.}
}

\author{
    \IEEEauthorblockN{Meng Wang$^{1,\dagger}$, Jintao Yang$^1$, Bin Yang$^{2,*}$, Hui Li$^{3,4}$, Tongxin Gong$^1$, Bo Yang$^{1,\dagger}$, Jiangtao Cui$^3$}
    \IEEEauthorblockA{
        \textit{$^1$School of Computer Science, Xi'an Polytechnic University, China}\\
        \textit{$^2$School of Data Science and Engineering, East China Normal University, China}\\
        \textit{$^3$School of Computer Science and Technology, Xidian University, China}\\
        \textit{$^4$Shanghai Yunxi Technology, China}\\
        wangmeng@xpu.edu.cn, byang@dase.ecnu.edu.cn, \{220721093, 220721091\}@stu.xpu.edu.cn,\\
        \{hli, cuijt\}@xidian.edu.cn, yangboo@stu.xjtu.edu.cn}
}

\maketitle

\begin{abstract}
    Patch-wise Transformer based time series forecasting achieves superior accuracy. However, this superiority relies heavily on intricate model design with massive parameters, rendering both training and inference expensive, thus preventing their deployments on edge devices with limited resources and low latency requirements. In addition, existing methods often work in an autoregressive manner, which take into account only historical values, but ignore valuable, easy-to-obtain context information, such as weather forecasts, date and time of day. To contend with the two limitations, we propose LiPFormer, a novel Lightweight Patch-wise Transformer with weak data enriching. First, to simplify the Transformer backbone, LiPFormer employs a novel lightweight cross-patch attention and a linear transformation-based attention to eliminate Layer Normalization and Feed Forward Network, two heavy components in existing Transformers. Second, we propose a lightweight, weak data enriching module to provide additional, valuable weak supervision to the training. 
    It enhances forecasting accuracy without significantly increasing model complexity as it does not involve expensive, human-labeling but using easily accessible context information. This facilitates the weak data enriching to plug-and-play on existing models. Extensive experiments on nine benchmark time series datasets demonstrate that LiPFormer outperforms state-of-the-art methods in accuracy, while significantly reducing parameter scale, training duration, and GPU memory usage. Deployment on an edge device reveals that LiPFormer only takes only 1/3 inference time compared to classic Transformers. In addition, we demonstrate that the weak data enriching can integrate seamlessly into various Transformer based models to enhance their accuracy, suggesting its generality.
\end{abstract}

\begin{IEEEkeywords}
    Time Series Data Forecasting, Weak Data Enriching, Lightweight, Patch-wise Transformer
\end{IEEEkeywords}

\section{Introduction}\label{sec:intro}
Time series constitutes a chronological sequence of data points that record successive states of an event. Forecasting is a fundamental task in time series data analysis, which aims to predict future values by tracking historical observations. Time series forecasting has received considerable research attention due to its crucial role in a spectrum of applications, such as finance~\cite{capistran2010multi}, weather~\cite{angryk2020multivariate}, energy~\cite{park1991electric}, and traffic~\cite{DBLP:journals/tkde/YangGY22,ding2018ultraman,chen2019real,chen2001freeway,fang2021mdtp,DBLP:conf/ijcai/YangGHT021,kieu2024Team}. Accurate forecasts can provide reliable data support, facilitating sound decision-making. For instance, industrial embedded sensors~\cite{ZhangLCZZ21} collect real-time operational data (e.g., temperature, pressure) from machinery to anticipate equipment health and failure risks. Power grids~\cite{LvWZGT22} utilize historical data from smart device (e.g., power loads, renewable energy production) to predict future power demand and supply, ensuring grid stability.

Time series forecasting has achieved remarkable advancement with a variety of architectures~\cite{DBLP:journals/pvldb/ChengCGZWYJ23,DBLP:journals/pacmmod/Wu0ZG0J23,DBLP:journals/vldb/WuWYZGQHSJ24,DBLP:journals/pvldb/QiuHZWDZGZJSY24,David2024Qcore,miao2024condensation,DBLP:journals/pvldb/ZhaoGCHZY23}, including the recurrent neural networks (RNNs)~\cite{elman1990finding,hochreiter1997long,salinasDeepARProbabilisticForecasting2020} and Transformers~\cite{vaswaniAttentionAllYou,DBLP:conf/iclr/ChenZ0SWW0G24}. Compared to RNNs, 
Transformers have revolutionized in both natural language processing (NLP)~\cite{radford2019language} and computer vision (CV)~\cite{liuSwinTransformerHierarchical2021} fields via their attention mechanism, enabling better understanding of global 
correlations. Benefiting from that, specialized Transformer-based architectures for time series forecasting emerged, e.g., Informer~\cite{zhouInformerEfficientTransformer2021}, Autoformer~\cite{wuAutoformerDecompositionTransformers}, FEDformer~\cite{zhouFEDformerFrequencyEnhanced2022}. These works 
attempt to preserve order information among time series data elements by Positional Encoding (PE). However, unlike natural language, the lack of semantics in numerical data makes PE invalid to capture sequential dependencies. 
Fortunately, the \textit{Patching} technique, inspired by a recent linear strategy (DLinear)~\cite{zengAreTransformersEffective}, segments time series data into subseries-level patches and assists Transformer-based models~\cite{nieTimeSeriesWorth2023,zhangCROSSFORMERTRANSFORMERUTILIZING2023,ekambaramTSMixerLightweightMLPMixer2023} in perceiving the order information.

Despite the progress made by patch-wise Transformers 
in time series forecasting, they face two substantial challenges.  

\textbf{Challenge 1: Intricate Models with Massive Parameters.} Heavyweight Transformer models, characterized by complicated modules and extensive parameters, incur prohibitive resource requirements and latency. The vanilla Transformer~\cite{vaswaniAttentionAllYou} exhibits $O(N^2)$ 
complexity ($N$ denotes time series length), rendering it unfriendly for 
training and deployment in resource-constrained scenarios. Rapid-response time series analysis tasks, increasingly prevalent in industrial~\cite{10187678} and networking~\cite{10361600} domains, are hindered by edge devices' limited computational power and memory to execute intricate algorithms, especially early and low-cost devices. Designing a tailored Transformer model that simultaneously achieves lightweight architecture and enhanced predictive performance is urgently required. The crucial issue lies in differentiating between components in Transformers effective for time series analysis and those of lesser importance. Developing strategies to optimize the former and simplify the latter is nontrivial.

\textbf{Challenge 2: Autoregressive forecasting without considering contexts.} Transformer-based 
models typically work in an autoregressive manner and predict time series solely on historical data, neglecting an explicit inductive bias: Future value changes (especially sudden changes) are highly correlated with future apriori contexts, like weather forecasts, date, or time of day. For instance, photovoltaic generation~\cite{CarriereVPK20} fluctuates due to weather variations, which are irrelevant to past electricity generation series data. Similarly, textual pollution severity labels (heavy, moderate, light) can provide weak supervision to aid in PM$_{2.5}$ prediction~\cite{YuHZGYL23}. Thus, incorporating easily obtainable weak labels as prior knowledge can enhance understanding of future time series dynamics. However, a unified multimodal framework is lacking in extracting features from explicit weak labels, such as textual (weather, wind direction, air pollution level, etc.) and numerical (temperature, humidity, etc.) covariates. Additionally, not all scenarios possess available explicit future covariates. While implicit weak labels (time of day, peak/non-rush hours, etc.) can be augmented via temporal feature encoding~\cite{zhouInformerEfficientTransformer2021,nieTimeSeriesWorth2023}, the absence of a decoder in patch-wise Transformer models hinders the availability of implicit future covariates for guiding predictions.

In order to explore the untapped potential of lightweight and weak labels in the Transformer architecture, this paper proposes a novel \textbf{\underline{Li}ghtweight \underline{P}atch-wise Trans\underline{Former} with weak data enriching} (\textbf{LiPFormer}) for time series data forecasting, which offers superior performance in both universal and resource-constrained environments. Specifically, an attention-based lightweight backbone network and a weakly supervised future covariate framework are devised to tackle the challenges above.

\textbf{Lightweight Patch-wise Transformer.} The lightweight backbone network, based on the multi-head self-attention mechanism, solves the first limitation through three key strategies as follows. 1) \textit{\textbf{Integrated cross-patch attention}}. We incorporate a newly designed cross-patch attention mechanism to the existing patching technique, simultaneously reducing complexity and improving predictive capability. Patching time series data dramatically reduces the number of input sequences to $O(N^2/pl^2)$ ($pl$ denotes patch length). The capability of patching mechanism~\cite{nieTimeSeriesWorth2023} to perceive local dependencies was inspired by DLinear. 
Motivated by trend components, another principle of DLinear, we extend to capture the global trend correlations across patches. By extracting fixed-position data points in each patch to construct trend sequences spanning all patches (details in Section~\ref{ssc:mc}), our patch-wise attention effectively substitutes Positional Encoding in the vanilla Transformer, perceiving both local and global sequential information. 2) \textit{\textbf{Layer Normalization elimination}}. Given the cross-patch's sampling manner to capture global trends, it partially supplants the generalization function of Layer Normalization (LN). Moreover, the uniform patch size of time series avoids the issue of varying token lengths characteristic of natural language, diminishing LN's utility~\cite{zerveas2021transformer}. Hence, its exclusion in our design is justified. 3) \textbf{FFNs-less (Feed Forward Networks-less)} linear attention. Recognizing the fundamental semantic disparities between textual and numerical sequences, we devise a novel FFNs-less attention scheme employing linear transformations. Transformers were originally tailored for language tasks and the nonlinear mappings of FFNs are more inclined towards learning semantic and syntactic structural information. DLinear's insights suggest that linear connections might suffice for time series representation. Thus, we employ a lightweight Multi-Layer Perceptron (MLP)
instead of the two-layer ascending and descending FFNs, significantly reducing the parameter scale from $O(8\times hd^2)$ to $O(hd\times pl)$ ($hd$ denotes hidden feature dimension). Encouragingly, deployment trials on an edge device with only CPU (with 16GB RAM, 6 cores, and 12 threads) reveal that the inference time of the lightweight LiPFormer is less than 1\% of that of Transformer models. 

\textbf{Weak Data Enriching}.
Typically, acquiring high-quality future covariate data are often economically pricey and labor-intensive. Fortunately, weakly supervised learning mitigates this issue by leveraging data augmentation to lower modeling barriers. Specifically, publicly accessible weather forecasts serve as expert annotations, while date-related contexts are augmented when explicit future features, like weather forecasts, are unavailable. The weak label enriching architecture branches into two policies depending on the availability of explicit future covariates. 1) \textit{\textbf{For scenarios with explicit weak label}}, there exist multimodal nature and cross-channel correlations. Textual and numerical future attributes are encoded into vectors with the same dimension, followed by an attention module to capture their dependencies. Similar to the backbone network, we employ a simplified MLP instead of the heavyweight Transformer for output, achieving a tradeoff between prediction accuracy and efficiency. 2) \textit{\textbf{In the absence of explicit covariates}}, temporal attributes (e.g., holidays, weekdays, rush hours) implicitly embraces meaningful semantic information. We encode these textual future covariates and embed them in a semantic space to maximize their correlation with the future ground truth 
time series (referred to as target sequences). 
We devise a contrastive learning framework for the augmented temporal features, adopting a dual encoder module—target sequences vs. future covariates—to effectively model their latent correlation. Notably, benefiting from the aligned dimension of target sequences and future covariates, the weak label enriching architecture can be seamlessly transplanted into existing time series forecasting frameworks 
and enhance their prediction capacities (detailed in Section~\ref{ssc:mc}).

Our main contributions in this paper are outlined as follows.

\begin{itemize}
    \item We propose LiPFormer, a novel patch-wise Transformer architecture with weak label enriching that lightweights the backbone network and integrates a weakly supervised architecture for future covariates, enhancing the predictive performance in time series forecasting.

    \item We present a novel weakly supervised architecture to learn the impact of future covariates. Based on data augmentation, a dual encoder contrastive learning framework is designed to uniformly model the correlation between textual, numerical, and implicit temporal future contexts.

    \item We innovate by integrating a cross-patch attention mechanism, eliminating Layer Normalization and Positional Encoding, and devising an FFNs-less linear attention to successfully simplify the Transformer architecture.

    \item Extensive evaluations on benchmark datasets demonstrate that LiPFormer significantly outperforms state-of-the-art methods in training speed, inference time, parameter size and accuracy. Deployment trials on a CPU-only edge device further confirm LiPFormer's lightweight superiority in resource-constrained scenarios over Transformer.

    \item Experiments on two real-world datasets with explicit future covariates validate the superiority of our weakly supervised architecture. Extension tests reveal it can be seamlessly transplanted into diverse time series forecasting models, contributing to improved performance.
\end{itemize}
\section{RELATED WORK}\label{sec:rl}

In this section, we overview existing 
time series forecasting models classified into four categories: Transformer-based, MLP-based, Deep Learning models, and patch-wise models.

\textbf{Transformers}:
RNNs were introduced to model temporal dependencies~\cite{ChungGCB14,wen2017multi,QinSCCJC17} 
for short-term series prediction. A recent study~\cite{YuHZGYL23} followed the general autoregressive manner to predict PM$_{2.5}$, leveraging easy-to-obtain categorical information to supervise the prediction. However, RNNs were susceptible to the vanishing gradient when handling long-term series. Based on self-attention mechanism, Transformer models~\cite{vaswaniAttentionAllYou} avoided the recurrent structure and were superior in capturing long-term dependencies. Nonetheless, attention mechanism suffers from high time and space complexity $O(N^2)$ and insensitivity to local context. LogTrans~\cite{SenYD19} 
proposed sparse convolutional self-attention to reduce complexity to $O(N(\log N)^2)$ incorporating local context into attention. Reformer~\cite{KitaevKL20} employed a locality-sensitive hashing attention, taking into account attention vectors only within the same hash buckets. Reversible residuals reduced complexity to $O(N \log N)$. Zhou et al.~\cite{zhouInformerEfficientTransformer2021} introduced ProbSparse self-attention, exploiting sparsity in attention parameters to
decrease the complexity. Conformer~\cite{li2023towards} reduces the complexity of the attention mechanism by $O(N)$ using window attention and a novel RNN network. Triformer~\cite{CirsteaG0KDP22}
leveraged a triangular structure and matrix factorization to achieve linear complexity. Inspired by stochastic processes, Wu et al.~\cite{wuAutoformerDecompositionTransformers} devised an autocorrelation mechanism, substituting the traditional
attention with a series-wise one.
Instead of optimizing attention, PatchTST~\cite{nieTimeSeriesWorth2023} and Crossformer~\cite{zhangCROSSFORMERTRANSFORMERUTILIZING2023} explored a different patching strategy to achieved efficiency gains. Recently, iTransformer~\cite{liu2024itransformerinvertedtransformerseffective} employed variate-wise attention to facilitate information exchange among variables.
Despite the above advancements in tackling the local-agnostics issue from various attention mechanisms or patching techniques, they still inadequately capture global sequential trend information,
which limited the performance of Transformer models~\cite{zengAreTransformersEffective}.

\textbf{MLPs}: Transformers had 
been the prevailing method for time series forecasting until a linear alternative challenged their supremacy. Zeng et al.~\cite{zengAreTransformersEffective} proposed a direct multi-step DLinear method, essentially an MLP, outperforming conventional Transformers. The 
authors decomposited time series into trend and seasonal components, which inspired numerous linear studies~\cite{ekambaramTSMixerLightweightMLPMixer2023,chenTSMixerAllMLPArchitecture2023,dasLongtermForecastingTiDE2023,0002Z0KGJ23,TFMRN}. MLP-Mixer~\cite{ekambaramTSMixerLightweightMLPMixer2023} effectively replaced
self-attention with an MLP and contained patch processing. Inspired by visual MLP mixers, Chen et al.~\cite{chenTSMixerAllMLPArchitecture2023} devised a two-stage framework with mixer layers and temporal projections. TiDE~\cite{dasLongtermForecastingTiDE2023} excelled under channel independence assumptions using a residual structure. Recently, a LightTS~\cite{0002Z0KGJ23} framework employed distillation and Pareto optimality techniques to substitute computationally intensive ensemble learning. TimeMixer~\cite{wangTIMEMIXERDECOMPOSABLEMULTISCALE2024} leveraged multi-resolution sequences in two Mixing modules for past and future feature extraction. While some of these models accounted for the impact of date-related implicit features on predictions, the lack of modeling future weak label (external covariates) limits their ability to exploit prior knowledge, which provides valuable supervision to enhance forecasting.

\textbf{Deep learning models}: FourierGNN~\cite{yiFourierGNNRethinkingMultivariate2023} innovatively treats time series values as graph nodes and performs predictions on hypervariable graphs. Deng et al.~\cite{SCNN} opted for an alternative policy, designing an SCNN network to individually model each component of the spatio-temporal patterns.

\textbf{Patching Models}: Most existing approaches directly utilized entire time series as model inputs, whereas PatchTST~\cite{nieTimeSeriesWorth2023} and TSMixer~\cite{ekambaramTSMixerLightweightMLPMixer2023} adopted a distinctive patching strategy. Patch-wise methods divide time series into subseries-level patches. Then it treats patches as input tokens to learn dependencies via attention mechanism. Data in a patch  
preserve local order information. However, as discussed earlier, patching attention lacks of global order awareness and the fixed patch size fails to accommodate different temporal scales, degrading model generalization.

Table~\ref{tab:relwork} summarizes representative time series forecasting methods, categorized according to their lightweight nature and consideration of future weak label. Our proposed LiPFormer bridges the gap in this research field.
\begin{table}[htbp]
    \caption{Representative time series forecasting models.}
    \label{tab:relwork}
    \centering
    \begin{tabular}{|c|c|c|}
        \hline
                                                                          & Heavyweight models                                                                                                                                                                                                                                                                                                            & Lightweight models                                                                                                                                                                                                                                                                                              \\ \hline
        \begin{tabular}[c]{@{}c@{}}Not consider\\ weak label\end{tabular} & \begin{tabular}[c]{@{}c@{}}RNNs~\cite{ChungGCB14,wen2017multi,QinSCCJC17}\\ Transformer~\cite{vaswaniAttentionAllYou}\\ Autoformer~\cite{wuAutoformerDecompositionTransformers}\\ LogTrans~\cite{SenYD19}\\ PatchTST~\cite{nieTimeSeriesWorth2023}\\ Crossformer~\cite{zhangCROSSFORMERTRANSFORMERUTILIZING2023}\\
        SCNN~\cite{SCNN}\\ Informer~\cite{zhouInformerEfficientTransformer2021}\end{tabular} & \begin{tabular}[c]{@{}c@{}}LightTS~\cite{0002Z0KGJ23}\\ Reformer~\cite{KitaevKL20}\\ Conformer~\cite{li2023towards}\\Triformer~\cite{CirsteaG0KDP22}\\ DLinear~\cite{zengAreTransformersEffective}\\ TSMixer~\cite{chenTSMixerAllMLPArchitecture2023} \end{tabular} \\ \hline
        \begin{tabular}[c]{@{}c@{}}Consider\\ weak label\end{tabular}     & \begin{tabular}[c]{@{}c@{}}CGF~\cite{YuHZGYL23}\\ TiDE~\cite{dasLongtermForecastingTiDE2023}\end{tabular}                                                                                                                                                                                                           & LiPFormer (Ours)                                                                                                                                                                                                                                                                                                \\ \hline
    \end{tabular}
\end{table}

\begin{figure*}
	\begin{minipage}{0.62\linewidth}
		\centering
		\includegraphics[width=\textwidth]{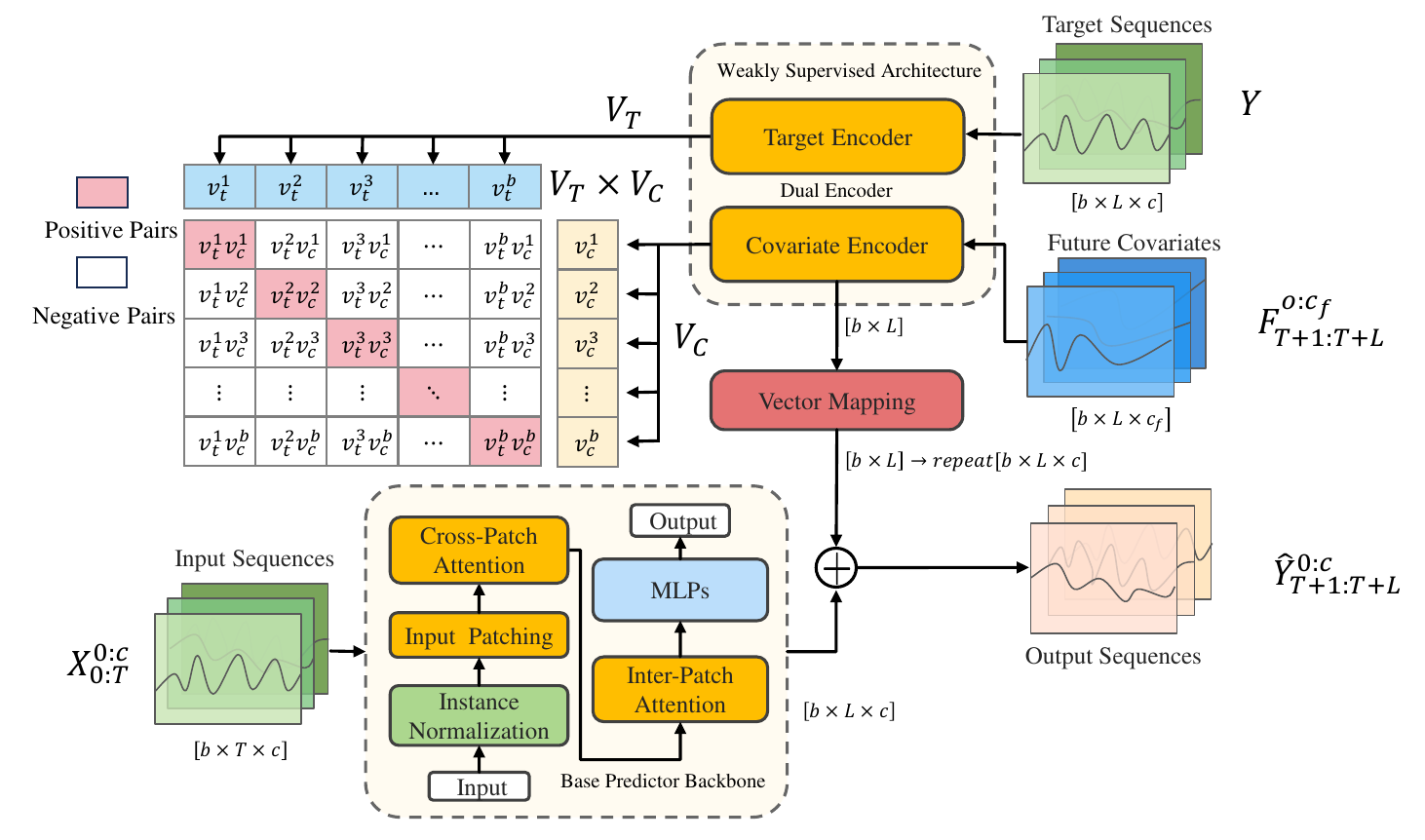}
		\caption{The architecture of LiPFormer. The Base Predictor backbone network 
			comprises two patch-wise attentions and simplified MLPs. The weakly supervised Dual Encoder is a contrastive learning architecture, consisting of a Covariate Encoder and a Target Encoder, to model the correlation between future attributes.}
		\label{fig:workflow}
	\end{minipage}
	\hspace{1ex}
	\begin{minipage}{0.36\linewidth}
		\centering
		\includegraphics[width=0.85\linewidth]{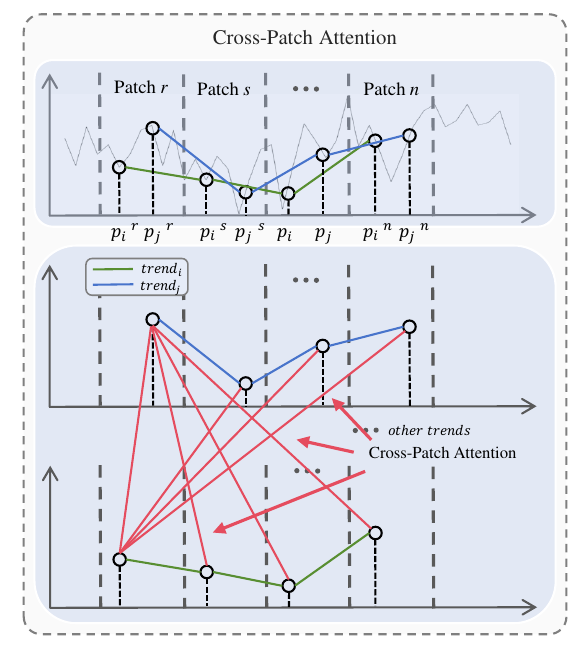}
		\vspace{-1ex}
		\caption{The construction of trend sequences and Cross-Patch attention.} 
	\label{fig:crossattn}
	\vspace{3ex}
	\centering
	\includegraphics[width=\linewidth]{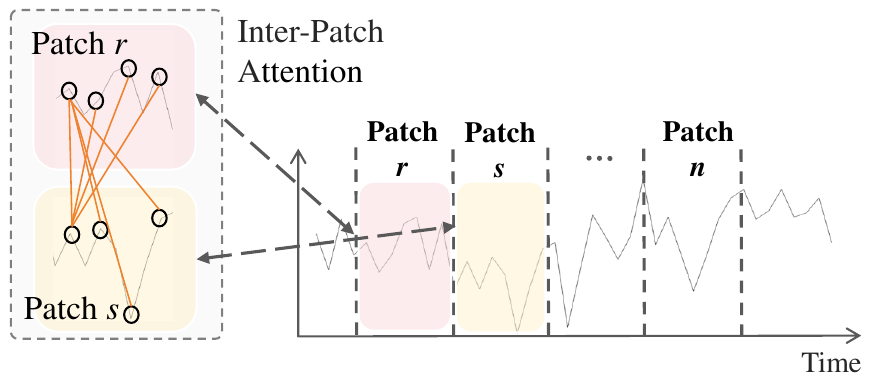}
	\vspace{-4.5ex}
	\caption{Patch division and Inter-Patch attention.}
	\label{fig:interattn}
	\end{minipage}
\end{figure*}

\section{METHODOLOGY} \label{se:meth}

In this section, we first define the notations 
and formally describe 
the time series forecasting task. Then we elaborate the training paradigm 
and our proposed lightweight Transformer architecture, following the order of data flow.

\subsection{Notations}
Frequently used notations in this paper are defined as follows.
$c$ and $c_f$: channels (features) of a mutivarite time series and the corresponding future covariates, they possibly are different;
$X^{0:c}_{0:T}$: an input time series (sequences) of length $T$ that consists of $T$ historical data points (a.k.a. time steps);
$F^{0:c_f}_{T+1:T+L}$: future covariates of length $L$;
$Y^{0:c}_{T+1:T+L}$ and $\hat{Y}^{0:c}_{T+1:T+L}$: the ground truth and forecast future sequence values of length $L$, respectively;
$\hat{Y}_{base}$: intermediate prediction results of Base Predictor, independent of future covariates;
$n$: number of patches;
$pl$: patch length;
$nt$: number of target patches, which is the ratio of predictive length divided by patch length;
$hd$: hidden feature dimensions;
$b$: batch size.

Given a historical time series $X^{0:c}_{0:T}$ and future covariates $F^{0:c_f}_{T+1:T+L}$, the multivariate forecasting task can be formally defined as the prediction of future values:
\begin{displaymath}
    \hat{Y}^{0:c}_{T+1:T+L} = \mathit{H}(X^{0:c}_{0:T},F^{0:c_f}_{T+1:T+L}),
\end{displaymath}
where $H$ denotes the proposed forecasting architecture.

\subsection{Training Methodologies} \label{se:tm}
LiPFormer involves two main training process, pre-training and prediction-oriented training. The weak label enriching part, depicted as Weakly Supervised Architecture in the top part of Figure~\ref{fig:workflow}, is a contrastive learning-based pre-training module with dual encoders that characterize target sequences (i.e., ground truth future time series) and weak labels, respectively. The backbone network, Base Predictor, performs prediction-oriented training over input sequences. 

Departing from prior representation learning methods that focus on encoding input sequences, we aim to extract representation vectors for easily accessible weak labels. As discussed in Section~\ref{sec:intro}, weak label enriching can provide additional and valuable weak supervision to the prediction. We leverage both explicit (textual and numerical external factors, e.g., weather, temperature) and implicit (temporal attributes, e.g., date, time of day) weak labels as future covariates. We follow a paradigm to regard future covariates as expert annotations, and utilize temporal information to augment weak data. Notably, the pre-training of weak labels serves to guide the Base Predictor in making predictions, instead of functioning as encoders to embed themselves as extra features of input time series.

Since patch-wise Transformers do not possess any decoder, in the absence of explicit weak labels, they cannot leverage future temporal features for guiding the Base Predictor. To this end, we devise the contrastive learning-based pre-training architecture (detailed in Section~\ref{ssse:rp}) to attain the implicit future feature representations via a joint encoding of ``covariate-target'' pair. Given a batch of $b$ covariate-target pairs, the ``covariate'' (resp., ``target'') element corresponds to a representative vector $\mathcal{V}_{c}^{(j)}$ (resp., $\mathcal{V}_{t}^{(i)}$) of future covariates $\mathcal{V_C}=\{\mathcal{V}_{c}^{(1)},\cdots,\mathcal{V}_{c}^{(b)}\}$ (resp., target sequences $\mathcal{V_T}=\{\mathcal{V}_{t}^{(1)},\cdots,\mathcal{V}_{t}^{(b)}\}$), where $\mathcal{V}_{c}^{(j)}$ (resp., $\mathcal{V}_{t}^{(i)}$) is of length $L$. This pre-training process is conducted upon learning the two representation vectors $\mathcal{V_C},\mathcal{V_T}$ to estimate which of the $b^2$ pairs actually occurred. Specifically, we train dual encoders, Covariate Encoder and Target Encoder, aiming at maximizing the cosine similarity of the $b$ diagonal positive pairs while minimizing the embedding of the remaining $b^2-b$ negative pairs. We use symmetric cross-entropy loss $\mathcal{L}_{sce}$ to optimize the cosine similarity score as follows:
\begin{displaymath}
    \mathcal{L}_{sce} = \frac{1}{2}(\mathcal{L}_{ce}(logits,labels)^{(0)}+\mathcal{L}_{ce}(logits,labels))^{(1)}),
\end{displaymath}

where $logits=(\mathcal{V_T}\times\mathcal{V_C})\cdot e^t\in\mathbb{R}^{b\times b}$ and $labels=(1,\cdots,b)$ are conceptually consistent with those in CLIP~\cite{radfordLearningTransferableVisual2021}. $\mathcal{L}_{ce}(p,q)=-\sum_x{p(x)\log(q(x))}$ denotes cross-entropy loss, and the superscript $(0)$ and $(1)$ in $\mathcal{L}_{sce}$ means to calculate $\mathcal{L}_{ce}(p,q)$ separately by row and column.

In the Base Predictor training process, the input time series are normalized and patched, then fed into the lightweight backbone network. The prediction head outputs the base prediction $\hat{Y}_{base}$, which is guided by the aforementioned future covariate dual encoders to get the future sequence prediction $\hat{Y}$. 
To achieve a balance between accuracy, convergence, and robustness during the backbone network training, we adopt Smooth $\mathcal{L}1$ loss ($\mathcal{L}1_{Smooth}$)~\cite{lin2023petformer}. When the error between the predicted and true values is small, $\mathcal{L}1_{Smooth}$ employs $\mathcal{L}2$ loss (Mean Squared Error, MSE) to maintain high accuracy, with its differentiable nature aiding convergence during training. Conversely, for larger errors, $\mathcal{L}1_{Smooth}$ uses $\mathcal{L}1$ loss (Mean Absolute Error, MAE), which is less sensitive to outliers, thus reducing the risk of gradient explosion. The hyperparameter $\beta$ in $\mathcal{L}1_{Smooth}$ is used to adjust the error threshold, providing learning flexibility for diverse time series datasets.
\begin{equation}\nonumber
    \mathcal{L}1_{Smooth}=
    \begin{cases}
        \frac{1}{2\beta}\left\|Y-\hat{Y}\right\|^2 & \quad if \left\|Y-\hat{Y}\right\|<\beta, \\
        \left\|Y-\hat{Y}\right\|-\frac{\beta}{2}   & \quad otherwise.
    \end{cases}
\end{equation}

\begin{figure*}[htbp]
    \centering
    \begin{minipage}{0.65\linewidth}
        \centering
        \includegraphics[width=0.9\linewidth]{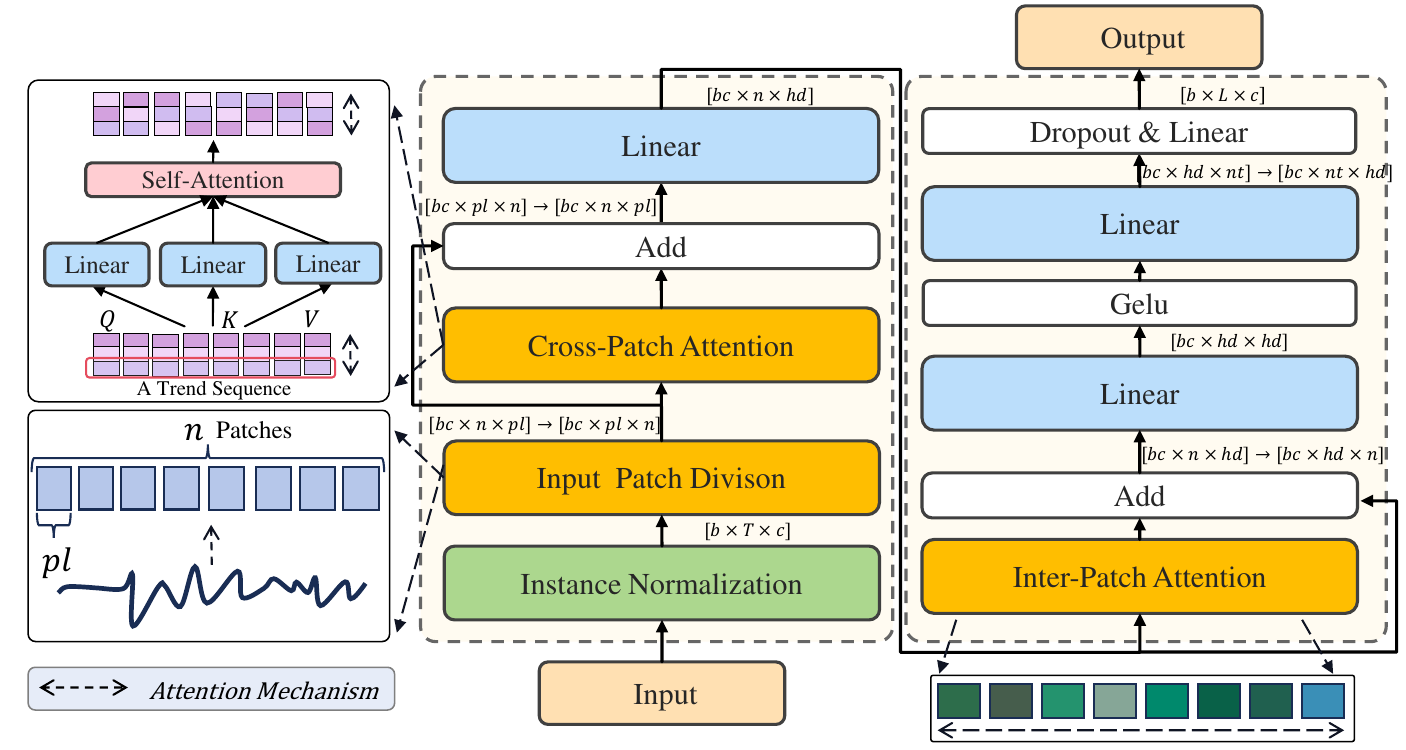}
        \caption{  The structure of Base Predictor block, which mainly comprises two patch-wise attentions and simplified MLPs.}
        \label{fig:modelarchitecture}
    \end{minipage}
    \qquad
    \begin{minipage}{0.3\linewidth}
        \centering
        \includegraphics[width=0.6\linewidth]{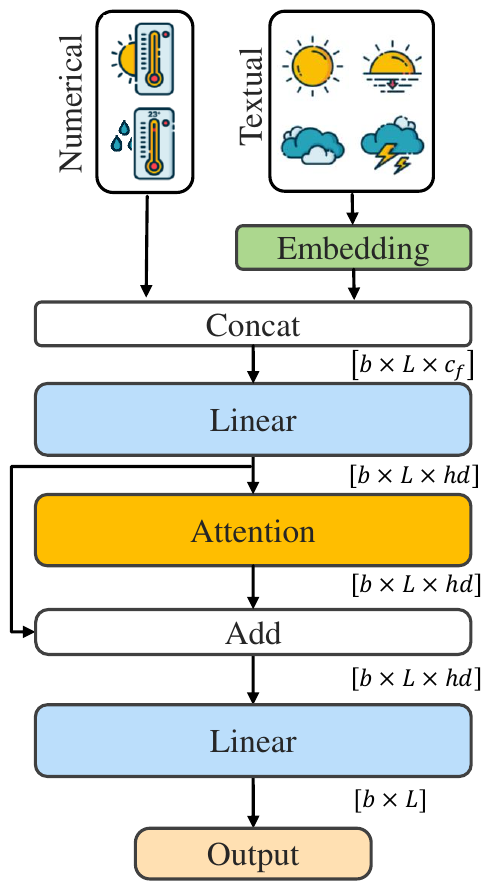}
        \caption{The structure of Covariate Encoder, which uses Res-attention and linear layers to model numerical and textual weak label supervision.}
        \label{fig:cov}
    \end{minipage}
\end{figure*}

\subsection{Model Components \label{ssc:mc}}

In this subsection, the first part delves into the intricate architecture of the Base Predictor, outlining its constituent components and functional mechanisms. In the second part (Section~\ref{ssse:rp}), we elaborate the details of Weakly Supervised dual encoder framework, 
specifically tailored for the purpose of learning meaningful representations of future weak labels. The entire end-to-end workflow, inclusive of the modular structures and data processing steps for each mini-batch, is depicted in Figure~\ref{fig:workflow}.

\subsubsection{\bfseries Base Predictor}\label{ssc:bp}

As illustrated in Figure~\ref{fig:modelarchitecture}, the backbone network of LiPFormer first eliminates LN, as the fixed patch size evades the issue with varying input sequence lengths. Secondly, adopting patching and channel independence preserves local order information and extends the receptive horizon to learn correlations between features and time series. Notably, existing patch-wise models commonly adhere to a fixed patch size, treating each patch as an individual token akin to text-oriented models. However, this rigidity in patch scales often hampers the capacity to effectively capture temporally varying periodic attributes inherent in time series. As a result, such models demonstrate limited adaptability to the task of time series prediction. This presents a significant challenge in holistically capturing information encapsulated within both individual patches and the broader sequence context, while maintaining a constant patch length. To overcome this issue, based on the patch division of input sequences, LiPFormer applies two novel patch-wise attention mechanisms, Cross-Patch and Inter-Patch, to comprehensively understand global and local sequential dependencies. Finally, in light of the inherent differences between numerical and textual sequences, the heavyweight FFNs module originally devised for understanding language information is replaced with two different single-layer MLPs, resulting in the base prediction $\hat{Y}_{base}$.

    {\bfseries Instance Normalization.}
Time series may encounter distribution shifts. To mitigate this issue, we employ a straightforward normalization approach~\cite{zengAreTransformersEffective} without significantly increasing model complexity. The initial input time series $x^i_{0:T}$ are subtracted by the last value $x^i_T$ from each element, resulting in a new input sequences $x^{i\prime}_{0:T}$, and subsequently re-adding it at the intermediate output $\hat{y}^{i\prime}_{T+1:T+L}$, denoted as:
\begin{displaymath}
    x^{i\prime}_{0:T} = x^i_{0:T} - x^i_T, \quad
    \hat{y}^i_{T+1:T+L} = \hat{y}^{i\prime}_{T+1:T+L} + x^i_T.
\end{displaymath}

{\bfseries Channel Independence and Patching.}
The use of multivariate time series data to simultaneously predict all variables may be an inductive bias for time series forecasting, as the interrelationships between variables can be captured. However, recent researches~\cite{nieTimeSeriesWorth2023,zengAreTransformersEffective,dasLongtermForecastingTiDE2023} have shown that this idea is not an optimal solution. The channel independent strategy maps only univariate sequences to future values without taking into account the mutual information of other variables. This design enables each univariate to independently process data while sharing a common weight space, enhancing local semantic capturing and extending the observation horizon.

Additionally, we divide each univariate sequence into patches of size $pl$. A mini-batch $X^{0:c}_{b\times T}$ of batch size $b$ will be reshaped into \(X^{0:c}_{b\times n \times pl}\), where \(n=\lceil T/pl\rceil\)
denotes the number of patches. With the patching technique, the number of input tokens (i.e., data units) can be reduced by a factor of $pl$ and the complexity dramatically drops to $O(N^2/pl^2)$, which significantly improves the model performance compared to the standard point-wise Transformer~\cite{nieTimeSeriesWorth2023}. On the other hand, data points in a patch are treated while preserving local coherence, thus exploiting local dependencies among them.

{\bfseries Novel Patch-wise Attention.}
The aforementioned patching manner logically treats each patch as a token. However, due to the variations in local granularity (e.g., periodicity) across diverse time series datasets, it is difficult to predetermine an appropriate patch size, limiting its generalization capability. Also, the patching mechanism does not intend to perceive the global order dependencies among tokens. To address these limitations, we introduce a novel \textbf{Cross-Patch} strategy.

We construct a \textit{global trend sequence} by arranging data points from a fixed position within patches in chronological order, as the construction of $trend_i$ and $trend_j$ shown in Figure~\ref{fig:crossattn}. According to the patch length, we can obtain a total of $pl$ trend sequences, each lagged by one time step. By applying attention across all these trend sequences, we can capture the mutual information at different positions of all the patches, perceiving sequential dependencies in the global trend. The lagging-like relationship among trend sequences is conducive to understanding the impact of historical information on future trends. Since data points in a trend sequence span beyond the scope of an individual patch, the generalization capability is less constrained by the fixed patch size, and multiple trend sequences collectively further enhance this generalization.

To explore correlations between patches, we map internal data points of patches onto an $hd$-dimensional latent feature vector using a MLP between them~\cite{ekambaramTSMixerLightweightMLPMixer2023,nieTimeSeriesWorth2023}, referred to as the \textbf{Inter-Patch} attention mechanism:
\begin{displaymath}
	X^{0:c}_{b\times n \times hd} = MLP(X^{0:c}_{b \times n \times pl}).
\end{displaymath}

Figure~\ref{fig:interattn} illustrates how an input time series is divided into patches and the proposed Inter-Patch attention works. For a particular patch $r$, the Inter-Patch attention mechanism focuses on the relevance between its data points to all counterparts within other patches, such as patch $s$. Similar to the self-attention mechanism, the Inter-Patch attention captures correlations between patch tokens.

Inspired by the mixer technique in CV domain~\cite{chenTSMixerAllMLPArchitecture2023}, the Cross-Patch attention performs feature mixing alongside the Inter-Patch attention with a MLP, enabling LiPFormer to learn correlations between local and trend features, thereby further alleviating degradation in generalization performance:
\begin{equation}
	x^{0:c}_{b \times n \times hd} = MLP(Attn(X^{0:c}_{b \times n \times pl})+X^{0:c}_{b \times n \times pl}),
\end{equation}
where $Attn$ denotes the self-attention mechanism of Vanilla Transformer~\cite{vaswaniAttentionAllYou}, mapping $X^i_{b\times n \times pl}$ to query matrices
\begin{math}
	\mathit{Q}^i = (x^i)^T\mathbf{W}^{\mathit{Q}},
\end{math} key matrices
\begin{math}
	\mathit{K}^i = (x^i)^T\mathbf{W}^{\mathit{K}},
\end{math}
and value matrices
\begin{math}
	\mathit{V}^i = (x^i)^T\mathbf{W}^{\mathit{V}},
\end{math}
with trainable parameter matrices $\mathbf{W}^{(\cdot)}$ and utilizing the $softmax$ function:
\begin{displaymath}
	Attn(X^i_{b \times n \times pl}) = Softmax(\frac{\mathit{Q}^i {\mathit{K}^i}^T}{\sqrt{d_k}})\mathit{V}^i.
\end{displaymath}
{\bfseries Elimination of Positional Encoding.} The positional information that represents the order of data points in time series is of great importance. As the self-attention mechanism cannot effectively retain the order information, some types of Positional Encoding (PE) were utilized by classical Informer~\cite{zhouInformerEfficientTransformer2021}, Autoformer~\cite{wuAutoformerDecompositionTransformers}, and FEDformer~\cite{zhouFEDformerFrequencyEnhanced2022}. For the sake of simplifying internal details, these PE approaches for enhancing the temporal order of time series inputs are uniformly represented by the trainable matrices $\mathbf{W}^{PE}$. Then the existing attention mechanism with Positional Encoding can be depicted as $Attn(x^{0:c}_{b\times n\times hd}+\mathbf{W}^{PE})$.

Unlike the traditional Transformers, by using the Inter-Patch attention, it is available to capture comprehensive positional information both at the point and patch levels without using any Positional Encoding method. As a result, we eliminate it between patches for direct inputs:
\begin{equation}
    Attn_{Inter-Patch} = Attn(x^{0:c}_{b \times n \times hd}).
\end{equation}

{\bfseries Lightweight Architecture.} The aforementioned patching technique reduces the time complexity of Transformer's attention mechanism to $O(N^2/{pl}^2)$, while
it is still computationally expensive. To this end, we explore the impact of Transformer's other auxiliary modules on time series modeling. At the beginning of this section, we clarified substituting Layer Normalization, mainly pertinent to token length variations in NLP tasks and with limited effect on time series forecasting~\cite{zerveas2021transformer}, with Instance Normalization. 
Our experiments demonstrate that LN tends to have no improvement in the accuracy of time series prediction and over-deep layers can lead to overfitting due to the applied channel independent strategy.

In addition, Feed Forward Networks (FFNs) in the Transformer architecture, which are used to perform nonlinear mapping to learn semantic and syntactic structural information, is overly heavy and not effective in capturing time series features. Similar to LN,
the FFNs module lacks specificity for numerical time series data and exhibits prohibitively expensive computational overhead. Hence,
we devise two linear transformation-based single-layer MLPs to capture linear dependencies. Specifically, we change the FFNs to directly predict $\hat{Y}_{base}$ instead of making residuals and then predicting through the linear layer. The specific shape changes are as follows (also shown in Figure \ref{fig:modelarchitecture}), where the arrowheads denote reshaping operations, and $b\cdot c$ indicates the single dimension size into which the batch size $b$ and the number of channels $c$ are reshaped:
\begin{displaymath}
    \begin{split}
        \hat{Y}_{base} = MLP(Attn_{wo/p}) & \in \mathbb{R}^{b\cdot c\times n \times hd} \\ &\rightarrow  \mathbb{R}^{b\cdot c \times hd \times nt}\rightarrow  \mathbb{R}^{b \times L \times c}
    \end{split}
\end{displaymath}
In summary, the theoretical analyses in this subsection and empirical studies below (detailed in Section~\ref{ssc:LA}) give us the confidence to eliminate PE, LN and FFNs layers from the standard and patch-wise Transformers.

\subsubsection{\bfseries Covariate Encoder} \label{ssse:rp}
Existing time series forecasting models, designed 
predominantly on publicly accessible datasets, typically neglect the influence of future 
weak labels on 
prediction, despite the intuitive and strong correlations between them. Bridging this gap necessitates a unified framework to handle datasets with and without future covariates. However, linear models are inadequate for capturing nonlinear relationships between time series and future covariates, which can be textual and numerical attributes.

To this end, we propose a nonlinear module for weak label enriching. For datasets inclusive of explicit future covariates, we employ a co-trained encoder alongside the Base Predictor. Conversely, for datasets lacking such future weak label covariates, we augment weak data with temporal feature encodings~\cite{zhouInformerEfficientTransformer2021} that align data dimension with future values in the latent space. Inspired by CLIP~\cite{radfordLearningTransferableVisual2021}, we formulate data pairs in the form of $(F_{T+1:T+L}^{0:c_f},Y_{T+1:T+L}^{0:c})$, where the two elements denote 
the feature encoding of future covariates and the target sequences, respectively.

In particular, we devise a simplified Transformer-based architecture, Covariate Encoder, for encoding future covariates (see Figure~\ref{fig:cov}). The encoder classifies input weak labels into two categories: numerical (e.g., temperature, humidity) and textual (e.g., wind direction, date, day of the week). We first encode textual weak labels into embeddings and then concatenate them with numerical labels, denoted as follows:
\begin{equation}
    F_{CatEmb}^{0:c_f} = Concat(Embed(F^{c_t}),F^{c_n}) \in \mathbb{R}^{b \times L \times c_f},
\end{equation}
where $Embed$ and $Concat$ are the corresponding operations, $c_n$ and $c_t$ are the numbers of numerical and textual channels, and $c_f=c_n+c_t$.

Next, the concatenated data are first fed into a linear MLP layer to map all channels to a hidden feature size of $hd$, followed by a fully flattened operation (denoted as $Flat$) over a residual self-attention mechanism. Finally, two encoded vectors (i.e., $\mathcal{V_C}$ and $\mathcal{V_T}$) of the same dimension are obtained by another MLP layer, and we make the dual encoders converge by the pre-training paradigm introduced in subsection~\ref{se:tm}. The Covariate Encoder works as below:
\begin{align}
    F_{MLP}^{c_f}      & =MLP(F_{CatEmb}^{0:c_f})\in\mathbb{R}^{b\times L\times hd},                 \\
    F_{FlatAttn}^{c_f} & =Flat(Attn(F_{MLP}^{c_f})+F_{MLP}^{c_f})\in\mathbb{R}^{b \times L\cdot hd}, \\
    F_{PreTrain}^{c_f} & =MLP(F_{FlatAttn}^{c_f})\in\mathbb{R}^{b\times L}.
\end{align}

Note that, the Target Encoder, designed for encoding target sequences, adopts an architecture akin to that of the Covariate Encoder. The Target Encoder dispenses with embedding and concatenation steps, necessitating merely the direct replacement of $F_{MLP}^{c_f}$ in Equation (5) with the following $F_{MLP}^{c}$:
\begin{equation}
    F_{MLP}^{c}=MLP(Y_{T+1:T+L}^{0:c})\in\mathbb{R}^{b\times L\times hd}.
\end{equation}

During prediction, we freeze the parameters of the Covariate Encoder and map the representation vectors to the same dimensions as the output of the Base Predictor through a learnable linear layer. As illustrated in Figure~\ref{fig:workflow}, the Vector Mapping linear layer is trained alongside the Base Predictor, enabling it to learn the relative contributions of the Covariate Encoder's and backbone network's parameters. By weighting these contributions, the pre-trained knowledge captured from weak labels helps to compensate for biases in the target sequences, thereby guiding the final prediction results.
\begin{equation}
    \hat{Y}= \hat{Y}_{base}+MLP({F_{PreTrain}^{c_f}} )\in\mathbb{R}^{b\times L\times c}
\end{equation}

\begin{table}[htbp]
    \centering
    \caption{The statistics of datasets.}
    \label{tab:sof}
    \addtolength{\tabcolsep}{-4pt}
    \scriptsize\resizebox{\linewidth}{!}{
        \begin{tabular}{cccccccccc}
            \hline
            Datasets    & ETTh1 & ETTh2 & ETTm1 & ETTm2 & Weather & Electricity & Traffic & Electri-Price & Cycle \\
            \hline
            Variables   & 7     & 7     & 7     & 7     & 21      & 321         & 862     & 40            & 22    \\
            Timestamps  & 17420 & 17420 & 69680 & 69680 & 52696   & 26304       & 17544   & 35808         & 21864 \\
            Split Ratio & 6:2:2 & 6:2:2 & 6:2:2 & 6:2:2 & 7:1:2   & 7:1:2       & 7:1:2   & 7:1:2         & 7:1:2 \\ \hline
        \end{tabular}
    }
\end{table}

\section{EXPERIMENTS} \label{se:exp}
This section presents experimental studies on the proposed LiPFormer compared with state-of-the-art (SOTA) models over a series of forecasting benchmarks. Performance comparisons between Transformer-based methods are conducted in terms of training time and running memory. Ablation studies on Layer Normalization and Feed Forward Network demonstrate the effectiveness of lightweight.

\subsection{Experimental Settings}

\subsubsection{\bfseries Datasets}We evaluate the performance of the proposed LiPFormer model on seven multivariate time series datasets, including Weather\footnote{https://www.bgc-jena.mpg.de/wetter/}, Traffic\footnote{https://pems.dot.ca.gov/}, Electricity\footnote{https://archive.ics.uci.edu/ml/datasets/ElectricityLoadDiagrams20112014}, and 4 ETT datasets\footnote{https://github.com/zhouhaoyi/ETDataset} (ETTh1, ETTh2, ETTm1, ETTm2). The statistics of these datasets are summarized in Table~\ref{tab:expriments}. These datasets have been extensively utilized in the literature~\cite{zengAreTransformersEffective,nieTimeSeriesWorth2023,zhouInformerEfficientTransformer2021} for benchmarking time series forecasting models and are publically available in~\cite{zengAreTransformersEffective}. We follow the same data loading parameters (i.e., train:validate:test split ratio) in~\cite{zengAreTransformersEffective}. To evaluate the impact of the dual encoder architecture for future covariates on prediction performance, this study employs two additional datasets with future features, namely Cycle and Electri-Price. The former dataset\footnote{https://github.com/nkullman/CSE512\_A3} records the bicycle counts per hour passing through the Seattle Fremont Bridge from October 2012 to March 2015, with future features mainly consisting of weather forecast information such as temperature, humidity, visibility, and wind strength. The latter dataset, proprietary business data\footnote{https://github.com/wangmeng-xpu/LiPFormer}, documents the real-time electricity market prices in a Chinese province every 15 minutes over one year from 2021 to 2022. Its future characteristics incorporate weather forecasts and features such as renewable energy generation and electricity load forecasts provided by the grid dispatching center. Table~\ref{tab:datasets_detials} presents detailed features of the two datasets.

\begin{table*}[]
    \centering
    \caption{Multivariate long-term time series forecasting results with LiPFormer. Electri-Price and Cycle are two datasets with future features. The best results are highlighted in bold, and the second-best results are underlined.}
    \label{tab:expriments}
    \normalsize
    \addtolength{\tabcolsep}{-3pt}
    \resizebox{\textwidth}{!}{
        \begin{tabular}{cc|ccc|ccc|ccc|ccc|ccc|ccc|ccc}
        	\hline
        	\multicolumn{2}{c|}{Models}                                                         & \multicolumn{3}{c|}{\textbf{LiPFormer}}                                                                                                         & \multicolumn{3}{c|}{iTransformer}                                                                                                                         & \multicolumn{3}{c|}{TimeMixer}                                                                                                                               & \multicolumn{3}{c|}{FGNN}                                                                                                                                 & \multicolumn{3}{c|}{PatchTST}                                                                                                                     & \multicolumn{3}{c|}{DLinear}                                                                                                                & \multicolumn{3}{c}{TiDE}                                                                                                                               \\
        	\multicolumn{2}{c|}{Metric}                                                         & MSE            & \multicolumn{1}{c|}{MAE}            & Efficiency                                                                               & MSE      & \multicolumn{1}{c|}{MAE}     & Efficiency                                                                                                      & MSE         & \multicolumn{1}{c|}{MAE}   & Efficiency                                                                                                        & MSE      & \multicolumn{1}{c|}{MAE}     & Efficiency                                                                                                      & MSE            & \multicolumn{1}{c|}{MAE}            & Efficiency                                                                                 & MSE         & \multicolumn{1}{c|}{MAE}         & Efficiency                                                                                 & MSE            & \multicolumn{1}{c|}{MAE}            & Efficiency                                                                                      \\ \hline
        	& 96  & \textbf{0.359} & \multicolumn{1}{c|}{\textbf{0.379}} &                                                                                          & 0.392    & \multicolumn{1}{c|}{0.423}   &                                                                                         & 0.385       & \multicolumn{1}{c|}{0.417} &                                                                                           & 0.647    & \multicolumn{1}{c|}{0.561}   &                                                                                         & {\ul 0.375}    & \multicolumn{1}{c|}{0.399}          &                                                                                            & {\ul 0.375} & \multicolumn{1}{c|}{0.399}       &                                                                                            & {\ul 0.375}    & \multicolumn{1}{c|}{{\ul 0.398}}    &                                                                                                 \\
        	& 192 & \textbf{0.404} & \multicolumn{1}{c|}{\textbf{0.405}} &                                                                                          & 0.437    & \multicolumn{1}{c|}{0.455}   &                                                                                         & 0.424       & \multicolumn{1}{c|}{0.443} &                                                                                           & 0.688    & \multicolumn{1}{c|}{0.583}   &                                                                                         & 0.414          & \multicolumn{1}{c|}{{\ul 0.421}}    &                                                                                            & {\ul 0.405} & \multicolumn{1}{c|}{0.436}       &                                                                                            & 0.412          & \multicolumn{1}{c|}{0.422}          &                                                                                                 \\
        	& 336 & 0.444          & \multicolumn{1}{c|}{\textbf{0.424}} &                                                                                          & 0.456    & \multicolumn{1}{c|}{0.469}   &                                                                                         & 0.456       & \multicolumn{1}{c|}{0.459} &                                                                                           & 0.704    & \multicolumn{1}{c|}{0.601}   &                                                                                         & \textbf{0.431} & \multicolumn{1}{c|}{0.436}          &                                                                                            & 0.439       & \multicolumn{1}{c|}{0.443}       &                                                                                            & {\ul 0.435}    & \multicolumn{1}{c|}{0.433}          &                                                                                                 \\
        	\multirow{-4}{*}{\rotatebox{90}{ETTh1}}                                                       & 720 & {\ul 0,450}    & \multicolumn{1}{c|}{\textbf{0.453}} & \multirow{-4}{*}{\begin{tabular}[c]{@{}c@{}}0.75s\\ 0.14s\\ 0.27G\\ 66K\end{tabular}}    & 0.553    & \multicolumn{1}{c|}{0.537}   & \multirow{-4}{*}{\begin{tabular}[c]{@{}c@{}}1.17s\\ 0.19s\\ 18.02G\\ 6.4M\end{tabular}} & 0.601       & \multicolumn{1}{c|}{0.559} & \multirow{-4}{*}{\begin{tabular}[c]{@{}c@{}}2.87s\\ 0.58s\\ 1.42T\\ 4.27M\end{tabular}}   & 0.772    & \multicolumn{1}{c|}{0.654}   & \multirow{-4}{*}{\begin{tabular}[c]{@{}c@{}}0.58s\\ 0.1s\\ 1.05G\\ 592K\end{tabular}}   & \textbf{0.449} & \multicolumn{1}{c|}{0.466}          & \multirow{-4}{*}{\begin{tabular}[c]{@{}c@{}}2.11s\\ 0.32s\\ 17.09G\\ 6.90M\end{tabular}}   & 0.472       & \multicolumn{1}{c|}{0.490}       & \multirow{-4}{*}{\begin{tabular}[c]{@{}c@{}}0.28s\\ 0.05s\\ 4.15M\\ 18.62K\end{tabular}}   & 0.454          & \multicolumn{1}{c|}{{\ul 0.465}}    & \multirow{-4}{*}{\begin{tabular}[c]{@{}c@{}}7.84s\\ 0.96s\\ 1.13G\\ 2.53M\end{tabular}}         \\ \hline
        	& 96  & \textbf{0.265} & \multicolumn{1}{c|}{\textbf{0.327}} &                                                                                          & 0.303    & \multicolumn{1}{c|}{0.364}   &                                                                                         & 0.296       & \multicolumn{1}{c|}{0.354} &                                                                                           & 0.479    & \multicolumn{1}{c|}{0.496}   &                                                                                         & 0.274          & \multicolumn{1}{c|}{{\ul 0.336}}    &                                                                                            & 0.289       & \multicolumn{1}{c|}{0.353}       &                                                                                            & {\ul 0.27}     & \multicolumn{1}{c|}{{\ul 0.336}}    &                                                                                                 \\
        	& 192 & {\ul 0.335}    & \multicolumn{1}{c|}{\textbf{0.374}} &                                                                                          & 0.409    & \multicolumn{1}{c|}{0.422}   &                                                                                         & 0.384       & \multicolumn{1}{c|}{0.415} &                                                                                           & 0.568    & \multicolumn{1}{c|}{0.540}   &                                                                                         & 0.339          & \multicolumn{1}{c|}{{\ul 0.379}}    &                                                                                            & 0.383       & \multicolumn{1}{c|}{0.418}       &                                                                                            & \textbf{0.332} & \multicolumn{1}{c|}{0.38}           &                                                                                                 \\
        	& 336 & 0.364          & \multicolumn{1}{c|}{{\ul 0.395}}    &                                                                                          & 0.440    & \multicolumn{1}{c|}{0.450}   &                                                                                         & 0.383       & \multicolumn{1}{c|}{0.423} &                                                                                           & 0.692    & \multicolumn{1}{c|}{0.604}   &                                                                                         & \textbf{0.331} & \multicolumn{1}{c|}{\textbf{0.380}} &                                                                                            & 0.480       & \multicolumn{1}{c|}{0.465}       &                                                                                            & {\ul 0.36}     & \multicolumn{1}{c|}{0.407}          &                                                                                                 \\
        	\multirow{-4}{*}{\rotatebox{90}{ETTh2} }                                                      & 720 & {\ul 0.392}    & \multicolumn{1}{c|}{{\ul 0.425}}    & \multirow{-4}{*}{\begin{tabular}[c]{@{}c@{}}0.80s\\ 0.14s\\ 0.27G\\ 66K\end{tabular}}    & 0.439    & \multicolumn{1}{c|}{0.468}   & \multirow{-4}{*}{\begin{tabular}[c]{@{}c@{}}1.04s\\ 0.19s\\ 18.02G\\ 6.4M\end{tabular}} & 0.399       & \multicolumn{1}{c|}{0.453} & \multirow{-4}{*}{\begin{tabular}[c]{@{}c@{}}2.42s\\ 0.59s\\ 1.42T\\ 4.27M\end{tabular}}   & 1.107    & \multicolumn{1}{c|}{0.774}   & \multirow{-4}{*}{\begin{tabular}[c]{@{}c@{}}0.6s\\ 0.15s\\ 1.05G\\ 592K\end{tabular}}   & \textbf{0.379} & \multicolumn{1}{c|}{\textbf{0.422}} & \multirow{-4}{*}{\begin{tabular}[c]{@{}c@{}}2.07s\\ 0.33s\\ 17.09G\\ 6.90M\end{tabular}}   & 0.605       & \multicolumn{1}{c|}{0.551}       & \multirow{-4}{*}{\begin{tabular}[c]{@{}c@{}}0.29s\\ 0.04s\\ 4.15M\\ 18.62K\end{tabular}}   & 0.419          & \multicolumn{1}{c|}{0.451}          & \multirow{-4}{*}{\begin{tabular}[c]{@{}c@{}}7.83s\\ 1.27s\\ 1.13G\\ 2.53M\end{tabular}}         \\ \hline
        	& 96  & {\ul 0.296}    & \multicolumn{1}{c|}{\textbf{0.338}} &                                                                                          & 0.318    & \multicolumn{1}{c|}{0.366}   &                                                                                         & 0.305       & \multicolumn{1}{c|}{0.358} &                                                                                           & 0.403    & \multicolumn{1}{c|}{0.427}   &                                                                                         & \textbf{0.290} & \multicolumn{1}{c|}{{\ul 0.342}}    &                                                                                            & 0.299       & \multicolumn{1}{c|}{0.343}       &                                                                                            & 0.306          & \multicolumn{1}{c|}{0.349}          &                                                                                                 \\
        	& 192 & 0.336          & \multicolumn{1}{c|}{\textbf{0.360}} &                                                                                          & 0.347    & \multicolumn{1}{c|}{0.387}   &                                                                                         & 0.343       & \multicolumn{1}{c|}{0.379} &                                                                                           & 0.426    & \multicolumn{1}{c|}{0.440}   &                                                                                         & \textbf{0.332} & \multicolumn{1}{c|}{0.369}          &                                                                                            & {\ul 0.335} & \multicolumn{1}{c|}{{\ul 0.365}} &                                                                                            & {\ul 0.335}    & \multicolumn{1}{c|}{0.366}          &                                                                                                 \\
        	& 336 & {\ul 0.365}    & \multicolumn{1}{c|}{\textbf{0.379}} &                                                                                          & 0.380    & \multicolumn{1}{c|}{0.405}   &                                                                                         & 0.371       & \multicolumn{1}{c|}{0.394} &                                                                                           & 0.450    & \multicolumn{1}{c|}{0.454}   &                                                                                         & 0.366          & \multicolumn{1}{c|}{0.453}          &                                                                                            & 0.369       & \multicolumn{1}{c|}{{\ul 0.386}} &                                                                                            & \textbf{0.364} & \multicolumn{1}{c|}{{\ul 0.384}}    &                                                                                                 \\
        	\multirow{-4}{*}{\rotatebox{90}{ETTm1} }                                                      & 720 & \textbf{0.408} & \multicolumn{1}{c|}{\textbf{0.413}} & \multirow{-4}{*}{\begin{tabular}[c]{@{}c@{}}3.1s\\ 0.55\\ 0.27G\\ 66K\end{tabular}}      & 0.436    & \multicolumn{1}{c|}{0.439}   & \multirow{-4}{*}{\begin{tabular}[c]{@{}c@{}}3.56s\\ 0.71s\\ 18.02G\\ 6.4M\end{tabular}} & 0.427       & \multicolumn{1}{c|}{0.423} & \multirow{-4}{*}{\begin{tabular}[c]{@{}c@{}}10.62s\\ 2.15s\\ 1.42T\\ 4.27M\end{tabular}}  & 0.498    & \multicolumn{1}{c|}{0.481}   & \multirow{-4}{*}{\begin{tabular}[c]{@{}c@{}}4.49s\\ 0.46s\\ 1.05G\\ 592K\end{tabular}}  & 0.420          & \multicolumn{1}{c|}{0.533}          & \multirow{-4}{*}{\begin{tabular}[c]{@{}c@{}}8.55s\\ 1.54s\\ 17.09G\\ 6.90M\end{tabular}}   & 0.425       & \multicolumn{1}{c|}{{\ul 0.421}} & \multirow{-4}{*}{\begin{tabular}[c]{@{}c@{}}1.12s\\ 0.20s\\ 4.15M\\ 18.62K\end{tabular}}   & {\ul 0.413}    & \multicolumn{1}{c|}{\textbf{0.413}} & \multirow{-4}{*}{\begin{tabular}[c]{@{}c@{}}32.51s\\ 4.64s\\ 1.13G\\ 2.53M\end{tabular}}        \\ \hline
        	& 96  & \textbf{0.160} & \multicolumn{1}{c|}{\textbf{0.244}} &                                                                                          & 0.180    & \multicolumn{1}{c|}{0.273}   &                                                                                         & 0.181       & \multicolumn{1}{c|}{0.270} &                                                                                           & 0.225    & \multicolumn{1}{c|}{0.322}   &                                                                                         & 0.165          & \multicolumn{1}{c|}{0.255}          &                                                                                            & 0.167       & \multicolumn{1}{c|}{0.260}       &                                                                                            & {\ul 0.161}    & \multicolumn{1}{c|}{{\ul 0.251}}    &                                                                                                 \\
        	& 192 & {\ul 0.217}    & \multicolumn{1}{c|}{\textbf{0.285}} &                                                                                          & 0.243    & \multicolumn{1}{c|}{0.315}   &                                                                                         & 0.239       & \multicolumn{1}{c|}{0.313} &                                                                                           & 0.296    & \multicolumn{1}{c|}{0.368}   &                                                                                         & 0.220          & \multicolumn{1}{c|}{0.292}          &                                                                                            & 0.224       & \multicolumn{1}{c|}{0.303}       &                                                                                            & \textbf{0.215} & \multicolumn{1}{c|}{{\ul 0.289}}    &                                                                                                 \\
        	& 336 & {\ul 0.273}    & \multicolumn{1}{c|}{\textbf{0.322}} &                                                                                          & 0.299    & \multicolumn{1}{c|}{0.352}   &                                                                                         & 0.289       & \multicolumn{1}{c|}{0.340} &                                                                                           & 0.423    & \multicolumn{1}{c|}{0.460}   &                                                                                         & 0.278          & \multicolumn{1}{c|}{0.329}          &                                                                                            & 0.281       & \multicolumn{1}{c|}{0.342}       &                                                                                            & \textbf{0.267} & \multicolumn{1}{c|}{{\ul 0.326}}    &                                                                                                 \\
        	\multirow{-4}{*}{\rotatebox{90}{ETTm2} }                                                      & 720 & \textbf{0.348} & \multicolumn{1}{c|}{\textbf{0.372}} & \multirow{-4}{*}{\begin{tabular}[c]{@{}c@{}}3.24s\\ 0.55\\ 0.27G\\ 66K\end{tabular}}     & 0.382    & \multicolumn{1}{c|}{0.405}   & \multirow{-4}{*}{\begin{tabular}[c]{@{}c@{}}2.89s\\ 0.74s\\ 18.02G\\ 6.4M\end{tabular}} & 0.452       & \multicolumn{1}{c|}{0.455} & \multirow{-4}{*}{\begin{tabular}[c]{@{}c@{}}9.45s\\ 2.07s\\ 1.42T\\ 4.27M\end{tabular}}   & 0.497    & \multicolumn{1}{c|}{0.493}   & \multirow{-4}{*}{\begin{tabular}[c]{@{}c@{}}2.74s\\ 0.56s\\ 1.05G\\ 592K\end{tabular}}  & 0.367          & \multicolumn{1}{c|}{0.385}          & \multirow{-4}{*}{\begin{tabular}[c]{@{}c@{}}7.68s\\ 2.05s\\ 17.09G\\ 6.90M\end{tabular}}   & 0.397       & \multicolumn{1}{c|}{0.421}       & \multirow{-4}{*}{\begin{tabular}[c]{@{}c@{}}1.21s\\ 0.21s\\ 4.15M\\ 18.62K\end{tabular}}   & {\ul 0.352}    & \multicolumn{1}{c|}{{\ul 0.383}}    & \multirow{-4}{*}{\begin{tabular}[c]{@{}c@{}}32.01s\\ 4.21s\\ 1.13G\\ 2.53M\end{tabular}}        \\ \hline
        	& 96  & {\ul 0.131}    & \multicolumn{1}{c|}{{\ul 0.224}}    &                                                                                          & 0.147    & \multicolumn{1}{c|}{0.249}   &                                                                                         & 0.134       & \multicolumn{1}{c|}{0.231} &                                                                                           & 0.211    & \multicolumn{1}{c|}{0.319}   &                                                                                         & \textbf{0.129} & \multicolumn{1}{c|}{\textbf{0.222}} &                                                                                            & 0.140       & \multicolumn{1}{c|}{0.237}       &                                                                                            & 0.132          & \multicolumn{1}{c|}{0.229}          &                                                                                                 \\
        	& 192 & \textbf{0.147} & \multicolumn{1}{c|}{\textbf{0.238}} &                                                                                          & 0.169    & \multicolumn{1}{c|}{0.271}   &                                                                                         & 0.339       & \multicolumn{1}{c|}{0.414} &                                                                                           & 0.226    & \multicolumn{1}{c|}{0.331}   &                                                                                         & \textbf{0.147} & \multicolumn{1}{c|}{{\ul 0.24}}     &                                                                                            & 0.153       & \multicolumn{1}{c|}{0.249}       &                                                                                            & \textbf{0.147} & \multicolumn{1}{c|}{0.243}          &                                                                                                 \\
        	& 336 & {\ul 0.163}    & \multicolumn{1}{c|}{\textbf{0.254}} &                                                                                          & 0.190    & \multicolumn{1}{c|}{0.292}   &                                                                                         & 0.280       & \multicolumn{1}{c|}{0.370} &                                                                                           & 0.242    & \multicolumn{1}{c|}{0.345}   &                                                                                         & {\ul 0.163}    & \multicolumn{1}{c|}{{\ul 0.259}}    &                                                                                            & 0.169       & \multicolumn{1}{c|}{0.267}       &                                                                                            & \textbf{0.161} & \multicolumn{1}{c|}{0.261}          &                                                                                                 \\
        	\multirow{-4}{*}{\rotatebox{90}{Electricity} }                                                & 720 & 0.199          & \multicolumn{1}{c|}{\textbf{0.285}} & \multirow{-4}{*}{\begin{tabular}[c]{@{}c@{}}5.12s\\ 1.01s\\ 2.94G\\ 66K\end{tabular}}    & 0.236    & \multicolumn{1}{c|}{0.329}   & \multirow{-4}{*}{\begin{tabular}[c]{@{}c@{}}4.24s\\ 0.97s\\ 66.56G\\ 6.4M\end{tabular}} & 0.557       & \multicolumn{1}{c|}{0.933} & \multirow{-4}{*}{\begin{tabular}[c]{@{}c@{}}-\\ -\\ 2.03T\\ 4.27M\end{tabular}}           & 0.274    & \multicolumn{1}{c|}{0.370}   & \multirow{-4}{*}{\begin{tabular}[c]{@{}c@{}}3.63s\\ 0.78s\\ 1.51G\\ 592K\end{tabular}}  & {\ul 0.197}    & \multicolumn{1}{c|}{{\ul 0.29}}     & \multirow{-4}{*}{\begin{tabular}[c]{@{}c@{}}16.19s\\ 2.20s\\ 195.96G\\ 6.90M\end{tabular}} & 0.203       & \multicolumn{1}{c|}{0.301}       & \multirow{-4}{*}{\begin{tabular}[c]{@{}c@{}}2.61s\\ 0.46s\\ 47.57M\\ 18.62K\end{tabular}}  & \textbf{0.196} & \multicolumn{1}{c|}{0.294}          & \multirow{-4}{*}{\begin{tabular}[c]{@{}c@{}}1452.56s\\ 255.30s\\ 250.86G\\ 34.10M\end{tabular}} \\ \hline
        	& 96  & 0.382          & \multicolumn{1}{c|}{\textbf{0.243}} &                                                                                          & 0.421    & \multicolumn{1}{c|}{0.318}   &                                                                                         & 0.385       & \multicolumn{1}{c|}{0.276} &                                                                                           & 0.721    & \multicolumn{1}{c|}{0.478}   &                                                                                         & {\ul 0.367}    & \multicolumn{1}{c|}{{\ul 0.251}}    &                                                                                            & 0.410       & \multicolumn{1}{c|}{0.282}       &                                                                                            & \textbf{0.336} & \multicolumn{1}{c|}{0.253}          &                                                                                                 \\
        	& 192 & 0.397          & \multicolumn{1}{c|}{\textbf{0.255}} &                                                                                          & 0.455    & \multicolumn{1}{c|}{0.340}   &                                                                                         & 0.394       & \multicolumn{1}{c|}{0.281} &                                                                                           & 0.777    & \multicolumn{1}{c|}{0.494}   &                                                                                         & {\ul 0.385}    & \multicolumn{1}{c|}{0.259}          &                                                                                            & 0.423       & \multicolumn{1}{c|}{0.287}       &                                                                                            & \textbf{0.346} & \multicolumn{1}{c|}{{\ul 0.257}}    &                                                                                                 \\
        	& 336 & 0.411          & \multicolumn{1}{c|}{\textbf{0.260}} &                                                                                          & 0.487    & \multicolumn{1}{c|}{0.359}   &                                                                                         & 0.413       & \multicolumn{1}{c|}{0.290} &                                                                                           & 0.813    & \multicolumn{1}{c|}{0.503}   &                                                                                         & 0.398          & \multicolumn{1}{c|}{{\ul 0.265}}    &                                                                                            & 0.436       & \multicolumn{1}{c|}{0.296}       &                                                                                            & \textbf{0.355} & \multicolumn{1}{c|}{\textbf{0.26}}  &                                                                                                 \\
        	\multirow{-4}{*}{\rotatebox{90}{Traffic} }                                                    & 720 & 0.451          & \multicolumn{1}{c|}{{\ul 0.281}}    & \multirow{-4}{*}{\begin{tabular}[c]{@{}c@{}}3.82s\\ 1.02s\\ 7.91G\\ 66K\end{tabular}}    & 0.555    & \multicolumn{1}{c|}{0.394}   & \multirow{-4}{*}{\begin{tabular}[c]{@{}c@{}}3.14s\\ 0.63s\\ 177G\\ 6.4M\end{tabular}}   & 0.452       & \multicolumn{1}{c|}{0.312} & \multirow{-4}{*}{-}                                                                       & 0.856    & \multicolumn{1}{c|}{0.517}   & \multirow{-4}{*}{\begin{tabular}[c]{@{}c@{}}-\\ -\\ 4.06G\\ 592K\end{tabular}}          & {\ul 0.434}    & \multicolumn{1}{c|}{0.287}          & \multirow{-4}{*}{\begin{tabular}[c]{@{}c@{}}9.84s\\ 1.69s\\ 526.22G\\ 6.90M\end{tabular}}  & 0.466       & \multicolumn{1}{c|}{0.315}       & \multirow{-4}{*}{\begin{tabular}[c]{@{}c@{}}1.58s\\ 0.29s\\ 127.76M\\ 18.62K\end{tabular}} & \textbf{0.386} & \multicolumn{1}{c|}{\textbf{0.273}} & \multirow{-4}{*}{\begin{tabular}[c]{@{}c@{}}2652.14s\\ 399.65s\\ 1.77T\\ 88.49M\end{tabular}}   \\ \hline
        	& 96  & \textbf{0.146} & \multicolumn{1}{c|}{\textbf{0.186}} &                                                                                          & 0.159    & \multicolumn{1}{c|}{0.211}   &                                                                                         & {\ul 0.151} & \multicolumn{1}{c|}{0.200} &                                                                                           & 0.166    & \multicolumn{1}{c|}{0.236}   &                                                                                         & 0.152          & \multicolumn{1}{c|}{{\ul 0.199}}    &                                                                                            & 0.176       & \multicolumn{1}{c|}{0.237}       &                                                                                            & 0.166          & \multicolumn{1}{c|}{0.222}          &                                                                                                 \\
        	& 192 & \textbf{0.189} & \multicolumn{1}{c|}{\textbf{0.230}} &                                                                                          & 0.202    & \multicolumn{1}{c|}{0.251}   &                                                                                         & 0.193       & \multicolumn{1}{c|}{0.242} &                                                                                           & 0.208    & \multicolumn{1}{c|}{0.274}   &                                                                                         & {\ul 0.197}    & \multicolumn{1}{c|}{{\ul 0.243}}    &                                                                                            & 0.22        & \multicolumn{1}{c|}{0.282}       &                                                                                            & 0.209          & \multicolumn{1}{c|}{0.263}          &                                                                                                 \\
        	& 336 & \textbf{0.244} & \multicolumn{1}{c|}{\textbf{0.277}} &                                                                                          & 0.256    & \multicolumn{1}{c|}{0.291}   &                                                                                         & 0.242       & \multicolumn{1}{c|}{0.281} &                                                                                           & 0.255    & \multicolumn{1}{c|}{0.311}   &                                                                                         & {\ul 0.249}    & \multicolumn{1}{c|}{{\ul 0.283}}    &                                                                                            & 0.265       & \multicolumn{1}{c|}{0.319}       &                                                                                            & 0.254          & \multicolumn{1}{c|}{0.301}          &                                                                                                 \\
        	\multirow{-4}{*}{\rotatebox{90}{Weather}}                                                     & 720 & \textbf{0.313} & \multicolumn{1}{c|}{\textbf{0.326}} & \multirow{-4}{*}{\begin{tabular}[c]{@{}c@{}}2.89s\\ 0.52s\\ 0.78G\\ 66K\end{tabular}}    & 0.323    & \multicolumn{1}{c|}{0.342}   & \multirow{-4}{*}{\begin{tabular}[c]{@{}c@{}}2.26s\\ 0.50s\\ 5.12G\\ 6.4M\end{tabular}}  & 0.319       & \multicolumn{1}{c|}{0.334} & \multirow{-4}{*}{\begin{tabular}[c]{@{}c@{}}7.17s\\ 1.57s\\ 532.70G\\ 4.27M\end{tabular}} & 0.314    & \multicolumn{1}{c|}{2.000}   & \multirow{-4}{*}{\begin{tabular}[c]{@{}c@{}}1.60s\\ 0.32s\\ 0.39G\\ 592K\end{tabular}}  & {\ul 0.320}    & \multicolumn{1}{c|}{{\ul 0.335}}    & \multirow{-4}{*}{\begin{tabular}[c]{@{}c@{}}4.01s\\ 0.67s\\ 51.27G\\ 6.90M\end{tabular}}   & 0.323       & \multicolumn{1}{c|}{0.362}       & \multirow{-4}{*}{\begin{tabular}[c]{@{}c@{}}0.49s\\ 0.11s\\ 12.45M\\ 18.62K\end{tabular}}  & \textbf{0.313} & \multicolumn{1}{c|}{0.34}           & \multirow{-4}{*}{\begin{tabular}[c]{@{}c@{}}35.62s\\ 7.79s\\ 6.17G\\ 3.93M\end{tabular}}        \\ \hline
        	& 96  & \textbf{0.486} & \multicolumn{1}{c|}{\textbf{0.424}} &                                                                                          & 0.677    & \multicolumn{1}{c|}{0.549}   &                                                                                         & 0.621       & \multicolumn{1}{c|}{0.531} &                                                                                           & 0.703    & \multicolumn{1}{c|}{0.552}   &                                                                                         & 0.635          & \multicolumn{1}{c|}{0.537}          &                                                                                            & {\ul 0.572} & \multicolumn{1}{c|}{{\ul 0.480}} &                                                                                            & 0.585          & \multicolumn{1}{c|}{{\ul 0.480}}    &                                                                                                 \\
        	& 192 & \textbf{0.528} & \multicolumn{1}{c|}{\textbf{0.443}} &                                                                                          & 0.749    & \multicolumn{1}{c|}{0.592}   &                                                                                         & 0.720       & \multicolumn{1}{c|}{0.586} &                                                                                           & 0.729    & \multicolumn{1}{c|}{0.563}   &                                                                                         & 0.697          & \multicolumn{1}{c|}{0.591}          &                                                                                            & 0.720       & \multicolumn{1}{c|}{{\ul 0.447}} &                                                                                            & {\ul 0.618}    & \multicolumn{1}{c|}{0.520}          &                                                                                                 \\
        	& 336 & \textbf{0.459} & \multicolumn{1}{c|}{\textbf{0.446}} &                                                                                          & 0.747    & \multicolumn{1}{c|}{0.591}   &                                                                                         & 0.704       & \multicolumn{1}{c|}{0.575} &                                                                                           & 0.723    & \multicolumn{1}{c|}{0.557}   &                                                                                         & 0.773          & \multicolumn{1}{c|}{0.615}          &                                                                                            & 0.651       & \multicolumn{1}{c|}{{\ul 0.533}} &                                                                                            & {\ul 0.643}    & \multicolumn{1}{c|}{0.542}          &                                                                                                 \\
        	\multirow{-4}{*}{\rotatebox{90}{\begin{tabular}[c]{@{}c@{}}Electricity\\ Price\end{tabular}}} & 720 & \textbf{0.495} & \multicolumn{1}{c|}{\textbf{0.467}} & \multirow{-4}{*}{\begin{tabular}[c]{@{}c@{}}2.54s\\ 0.50s\\ 8.31G\\ 74.86K\end{tabular}} & 0.797    & \multicolumn{1}{c|}{0.648}   & \multirow{-4}{*}{\begin{tabular}[c]{@{}c@{}}1.63s\\ 0.31s\\ 1.22G\\ 6.4M\end{tabular}}  & 0.675       & \multicolumn{1}{c|}{0.566} & \multirow{-4}{*}{\begin{tabular}[c]{@{}c@{}}4.52s\\ 0.96s\\ 12.68G\\ 4.27M\end{tabular}}  & 0.696    & \multicolumn{1}{c|}{0.545}   & \multirow{-4}{*}{\begin{tabular}[c]{@{}c@{}}1.31s\\ 0.21s\\ 11.79M\\ 592K\end{tabular}} & 0.832          & \multicolumn{1}{c|}{0.629}          & \multirow{-4}{*}{\begin{tabular}[c]{@{}c@{}}2.74s\\ 0.50s\\ 4.88G\\ 6.90M\end{tabular}}    & 0.860       & \multicolumn{1}{c|}{{\ul 0.520}} & \multirow{-4}{*}{\begin{tabular}[c]{@{}c@{}}0.38s\\ 0.11s\\ 1.18M\\ 18.62K\end{tabular}}   & {\ul 0.632}    & \multicolumn{1}{c|}{0.538}          & \multirow{-4}{*}{\begin{tabular}[c]{@{}c@{}}3.28s\\ 0.65s\\ 0.22G\\ 2.02M\end{tabular}}         \\ \hline
        	& 96  & \textbf{0.136} & \multicolumn{1}{c|}{\textbf{0.221}} &                                                                                          & 0.182    & \multicolumn{1}{c|}{0.278}   &                                                                                         & 0.169       & \multicolumn{1}{c|}{0.265} &                                                                                           & 0.441    & \multicolumn{1}{c|}{0.464}   &                                                                                         & 0.160          & \multicolumn{1}{c|}{0.248}          &                                                                                            & 0.174       & \multicolumn{1}{c|}{0.254}       &                                                                                            & {\ul 0.150}    & \multicolumn{1}{c|}{{\ul 0.236}}    &                                                                                                 \\
        	& 192 & \textbf{0.145} & \multicolumn{1}{c|}{\textbf{0.230}} &                                                                                          & 0.212    & \multicolumn{1}{c|}{0.302}   &                                                                                         & 0.203       & \multicolumn{1}{c|}{0.303} &                                                                                           & 0.469    & \multicolumn{1}{c|}{0.475}   &                                                                                         & 0.167          & \multicolumn{1}{c|}{0.254}          &                                                                                            & 0.177       & \multicolumn{1}{c|}{0.254}       &                                                                                            & {\ul 0.158}    & \multicolumn{1}{c|}{{\ul 0.240}}    &                                                                                                 \\
        	& 336 & \textbf{0.152} & \multicolumn{1}{c|}{\textbf{0.235}} &                                                                                          & 0.232    & \multicolumn{1}{c|}{0.318}   &                                                                                         & 0.146       & \multicolumn{1}{c|}{0.240} &                                                                                           & 0.469    & \multicolumn{1}{c|}{0.472}   &                                                                                         & 0.179          & \multicolumn{1}{c|}{0.263}          &                                                                                            & 0.184       & \multicolumn{1}{c|}{0.256}       &                                                                                            & {\ul 0.167}    & \multicolumn{1}{c|}{{\ul 0.246}}    &                                                                                                 \\
        	\multirow{-4}{*}{\rotatebox{90}{Cycle}}                                                       & 720 & \textbf{0.159} & \multicolumn{1}{c|}{\textbf{0.236}} & \multirow{-4}{*}{\begin{tabular}[c]{@{}c@{}}1.09s\\ 0.19s\\ 0.84G\\ 74K\end{tabular}}    & 0.258    & \multicolumn{1}{c|}{0.337}   & \multirow{-4}{*}{\begin{tabular}[c]{@{}c@{}}1.05s\\ 0.18s\\ 1.02G\\ 6.4M\end{tabular}}  & 0.175       & \multicolumn{1}{c|}{0.263} & \multirow{-4}{*}{\begin{tabular}[c]{@{}c@{}}2.79s\\ 0.58s\\ 6.34G\\ 4.27M\end{tabular}}   & 0.482    & \multicolumn{1}{c|}{0.480}   & \multirow{-4}{*}{\begin{tabular}[c]{@{}c@{}}1.01s\\ 0.12s\\ 18M\\ 592K\end{tabular}}    & 0.214          & \multicolumn{1}{c|}{0.284}          & \multirow{-4}{*}{\begin{tabular}[c]{@{}c@{}}1.60s\\ 0.29s\\ 2.44G\\ 6.90M\end{tabular}}    & 0.192       & \multicolumn{1}{c|}{0.267}       & \multirow{-4}{*}{\begin{tabular}[c]{@{}c@{}}0.22s\\ 0.05s\\ 0.59M\\ 18.62K\end{tabular}}   & {\ul 0.181}    & \multicolumn{1}{c|}{{\ul 0.256}}    & \multirow{-4}{*}{\begin{tabular}[c]{@{}c@{}}1.70s\\ 0.21s\\ 0.10G\\ 1.92M\end{tabular}}         \\ \hline
        	Count                                                                         &     & \multicolumn{2}{c|}{\textbf{51}/{\ul 12}}            &                                                                                          & \multicolumn{2}{c|}{\textbf{0}/{\ul 0}} &                                                                                                                 & \multicolumn{2}{c|}{\textbf{0}/{\ul 1}}  &                                                                                                                   & \multicolumn{2}{c|}{\textbf{0}/{\ul 0}} &                                                                                                                 & \multicolumn{2}{c|}{\textbf{11}/{\ul 23}}            &                                                                                            & \multicolumn{2}{c|}{\textbf{0}/{\ul 11}}       &                                                                                            & \multicolumn{2}{c|}{\textbf{15}/{\ul 29}}            &                                                                                                 \\ \hline
        \end{tabular}
        }
    \begin{tablenotes}
        \item[1] \scriptsize{For the \textit{Efficiency} columns, values are exhibited, in order, as training time (seconds per epoch), inference time (seconds per inference), MACs and the number of model parameters (``-'' indicates insufficient GPU memory under the current experiment environment). These values correspond to the forecast sequence length of 96 and exhibit quantitative similarity for other lengths. The best and second-best counts are in the last row.}
    \end{tablenotes}
\end{table*}

\subsubsection{\bfseries Data \& Model Configuration} By default, we use the following data and model configurations: Input Sequence length $T=720$ , Patch length $pl=48$, Batch size $b=256$ , Forecast sequence length $L=\{96, 192, 336, 720\}$,  Hidden feature size $hd=512$, and Dropout = 0.5. LiPFormer utilizes the AdamW optimizer\cite{LoshchilovH19}. Training is performed in a distributed fashion with 1 GPU, 1 CPU and 32 GB of memory. For ETT datasets, we use a lower hardware and model configuration with high dropout to avoid overfitting, as the dataset is relatively small.  Training is performed with 10 epochs and used early stopping with 3 patients. We choose the final model based on the best validation score. For patching, we use a patch length $pl=48$ that divides the input length $T$ equally and without overlap. Every experiment is executed with the same random seed and the mean scores are reported. We use MSE $= \frac{1}{T}\sum_{i=0}^{T}{(\hat{X}_i-X_i)}^2$ and MAE $= \frac{1}{T}\sum_{i=0}^{T}{|\hat{X}_i-X_i|}$ to evaluate accuracy.

In addition, we also examine the training/inference time, the number of parameters and MACs (Multiply–Accumulate Operations) of the models. We conduct the experiments using the same input and output lengths of 96. A batch size of 32 is utilized for all datasets. Given the substantial number of channels and data volume present in both the Traffic and Electricity datasets, we employ a batch size of 8 for experiments on the two datasets, in order to better adapt to the actual GPU memory size. 

\subsubsection{\bfseries Baselines}We evaluated the following SOTA benchmarks: iTransformer~\cite{liu2024itransformerinvertedtransformerseffective}, TimeMixer~\cite{wangTIMEMIXERDECOMPOSABLEMULTISCALE2024}, FGNN~\cite{yiFourierGNNRethinkingMultivariate2023}, PatchTST~\cite{nieTimeSeriesWorth2023}, Dlinear~\cite{zengAreTransformersEffective}, and TiDE~\cite{dasLongtermForecastingTiDE2023}.

\subsection{Non-covariate Prediction}

\subsubsection{\bfseries Data Preprocess}Since these public datasets do not have future covariates, we can only construct features as the hour of the day, day of the week, day of the month, and month of the year as inputs for future covariates in a similar way to the time encoding in Informer.

\subsubsection{\bfseries Accuracy Improvements}In Table~\ref{tab:expriments}, we compare the accuracy of LiPFormer to the SOTA benchmarks. Since similar relative patterns can be observed in both MSE and MAE, we use the MAE metrics to interpret all results in this paper. LiPFormer significantly outperforms the existing benchmarks (DLinear: 10.4\%, iTransformer: 18\%, TimeMixer: 21\%, FGNN: 62\%). PatchTST and TiDE are the two strongest baseline models, while LiPFormer outperforms them by a narrow margin of 8.3\% and 5.4\%.

LiPFormer demonstrates strong multivariate long-term time series prediction performance, achieving top-two rankings in 64 out of 72 metrics across various scenarios, with 51 instances of first place. This convincingly surpasses other baselines, including the SOTA Transformer-based model, PatchTST. LiPFormer excels particularly on the ETT and Weather datasets, consistently outperforming SOTA on nearly all metrics. On larger datasets like electricity and transportation, where channel counts exceed 300 and 800, respectively, LiPFormer's performance marginally declines. This can potentially be attributed to its relatively smaller model capacity. Nonetheless, even in these challenging scenarios, LiPFormer maintains competitiveness or demonstrates superiority relative to alternative models. We believe that this is due to the fact that Cross-Patch attention and Inter-Patch attention better capture the continuity and global nature of the time series. At the same time, since the extra component of transformer does not significantly improve the accuracy, it leads us to a win-win situation in terms of accuracy and efficiency. In particular, for more volatile datasets (ETTh1, Weather), we prefer to use LiPFormer as a predictive model.

\subsection{Forecasting via Future Covariate}
In this part, we investigate the effectiveness of the Covariate Encoder framework using two datasets containing future weak labels, where the Electricity Price and Cycle datasets respectively include 61 and 22 covariates. 
The Electricity-Price dataset records the variations of electricity spot market from Shanxi province in China, and the Cycle dataset corresponds to the bicycle-riding scenario in Seattle, U.S. Their future weak labels are detailed in Table~\ref{tab:datasets_detials}.

\begin{table}[]
    \centering
    \caption{Details of the two datasets with future features.}
    \label{tab:datasets_detials}
    \addtolength{\tabcolsep}{-3pt}
    \resizebox{\linewidth}{!}{\begin{tabular}{c|c|c|c}
            \hline
            Datasets                                                                      & Future Covariates                              & \# of Fields & Data Field Type \\
            \hline
            \multirow{10}{*}{\begin{tabular}[c]{@{}c@{}}Electricity\\ Price\end{tabular}} & Unified Load Forecast (MW)                     & 1            & numerical       \\
                                                                                          & Outgoing forecast (MW)                         & 1            & numerical       \\
                                                                                          & Sum of wind and light projections              & 1            & numerical       \\
                                                                                          & Wind power projections                         & 1            & numerical       \\
                                                                                          & Photovoltaic Forecast                          & 1            & numerical       \\
                                                                                          & Max and Min temperature at a location          & 22           & numerical       \\
                                                                                          & Wind rating at a location                      & 11           & numerical       \\
                                                                                          & The direction of the wind at a location        & 11           & numerical       \\
                                                                                          & Weather conditions at a location               & 11           & categorical     \\
                                                                                          & holiday                                        & 1            & categorical     \\
            \hline
            \multirow{9}{*}{Cycle}                                                        & Max,Min,Mean temperature                       & 3            & numerical       \\
                                                                                          & Max,Min,Mean dewpointF                         & 3            & numerical       \\
                                                                                          & Max,Min,Mean humidity                          & 3            & numerical       \\
                                                                                          & Max,Min,Mean sea level pressure in             & 3            & numerical       \\
                                                                                          & Max,Min,Mean visibility miles                  & 3            & numerical       \\
                                                                                          & Max,Mean wind speed MPH and wind direct degree & 3            & numerical       \\
                                                                                          & Max gust speed MPH                             & 1            & numerical       \\
                                                                                          & precipitation in                               & 1            & numerical       \\
                                                                                          & cloud cover                                    & 1            & numerical       \\
                                                                                          & weekend                                        & 1            & categorical     \\\hline
        \end{tabular}}

\end{table}

As shown in the last two datasets of Table~\ref{tab:expriments}, the proposed LiPFormer surpasses all other comparative models in prediction accuracy, indicating that leveraging the weak supervision of future covariates through pre-training can effectively guide the predictions. It is worth noting that the TiDE model also took into account external factors for prediction, enabling it to outperform other counterparts, except for LiPFormer, on these two datasets. This further highlights the utility of weak labels, and also validates our claim that future value changes are highly correlated with apriori contexts. Compared to TiDE, LiPFormer's average MSE using future covariates is reduced by 20.6\% and 18.6\% on the two datasets. This is attributed to our model's ability to characterize the alignment of time series data achieved during the pre-training of covariates, demonstrating the superiority of the dual encoder strategy.

Furthermore, for datasets that do not contain explicit weak labels, augmenting the weak data with implicit temporal features can still enrich data understanding and significantly enhance prediction performance. As illustrated in Table~\ref{tab:implicit}, we examine forecast results on four datasets (with prediction lengths of 96) that lack explicit future covariates. By comparing the prediction outcomes with and without leveraging implicit temporal features for weak data enriching, it is evident that adopting the pre-trained model yields substantially positive results.

\begin{table*}
	\centering
	\caption{Univariate long-term time series forecasting results with LiPFormer. The best results are highlighted in bold, and the second-best results are underlined.}
	\label{tab:expuni}
	\normalsize
	\scriptsize\resizebox{0.9\textwidth}{!}{
		\begin{tabular}{cc|cccccccccc|cccc}
			\hline
			Models                   &      & \multicolumn{2}{c|}{LiPFormer}                       & \multicolumn{2}{c|}{iTransformer}                                            & \multicolumn{2}{c|}{TimeMixer}                                         & \multicolumn{2}{c|}{FGNN}                                  & \multicolumn{2}{c|}{PatchTST}            & \multicolumn{2}{c|}{DLinear}                         & \multicolumn{2}{c}{TiDE}               \\
			Metric                   &      & MSE            & \multicolumn{1}{c|}{MAE}            & MSE            & \multicolumn{1}{c|}{MAE}            & MSE         & \multicolumn{1}{c|}{MAE}         & MSE   & \multicolumn{1}{c|}{MAE}   & MSE                 & MAE                & MSE            & \multicolumn{1}{c|}{MAE}            & MSE                & MAE               \\ \hline
			& 96   & {\ul 0.057}    & \multicolumn{1}{c|}{\textbf{0.18}}  & 0.059          & \multicolumn{1}{c|}{0.189}          & 0.103       & \multicolumn{1}{c|}{0.252}       & 0.106 & \multicolumn{1}{c|}{0.256} & 0.059               & {\ul 0.189}        & \textbf{0.056} & \multicolumn{1}{c|}{\textbf{0.18}}  & 0.067              & 0.204             \\
			& 192  & {\ul 0.070}    & \multicolumn{1}{c|}{{\ul 0.209}}    & \textbf{0.068} & \multicolumn{1}{c|}{\textbf{0.204}} & 0.113       & \multicolumn{1}{c|}{0.270}       & 0.112 & \multicolumn{1}{c|}{0.265} & 0.074               & 0.215              & 0.071          & \multicolumn{1}{c|}{\textbf{0.204}} & 0.081              & 0.225             \\
			& 336  & \textbf{0.075} & \multicolumn{1}{c|}{\textbf{0.216}} & 0.078          & \multicolumn{1}{c|}{0.221}          & 0.102       & \multicolumn{1}{c|}{0.257}       & 0.126 & \multicolumn{1}{c|}{0.284} & {\ul 0.076}         & {\ul 0.22}         & 0.098          & \multicolumn{1}{c|}{0.244}          & 0.090              & 0.240             \\
			\multirow{-4}{*}{\rotatebox{90}{ETTh1}} & 720  & 0.102          & \multicolumn{1}{c|}{0.252}          & \textbf{0.080} & \multicolumn{1}{c|}{\textbf{0.230}} & 0.122       & \multicolumn{1}{c|}{0.288}       & 0.252 & \multicolumn{1}{c|}{0.423} & {\ul 0.087}         & {\ul 0.236}        & 0.189          & \multicolumn{1}{c|}{0.359}          & 0.104              & 0.254             \\ \hline
			& 96   & {\ul 0.134}    & \multicolumn{1}{c|}{{\ul 0.282}}    & 0.147          & \multicolumn{1}{c|}{0.307}          & 0.158       & \multicolumn{1}{c|}{0.316}       & 0.224 & \multicolumn{1}{c|}{0.384} & \textbf{0.131}      & 0.284              & \textbf{0.131} & \multicolumn{1}{c|}{\textbf{0.279}} & 0.164              & 0.320             \\
			& 192  & \textbf{0.157} & \multicolumn{1}{c|}{\textbf{0.314}} & {\ul 0.155}    & \multicolumn{1}{c|}{{\ul 0.320}}    & 0.205       & \multicolumn{1}{c|}{0.357}       & 0.248 & \multicolumn{1}{c|}{0.407} & 0.171               & 0.329              & 0.176          & \multicolumn{1}{c|}{{\ul 0.329}}    & 0.186              & 0.345             \\
			& 336  & {\ul 0.169}    & \multicolumn{1}{c|}{{\ul 0.330}}    & \textbf{0.159} & \multicolumn{1}{c|}{\textbf{0.326}} & 0.200       & \multicolumn{1}{c|}{0.359}       & 0.303 & \multicolumn{1}{c|}{0.450} & 0.171               & 0.336              & 0.209          & \multicolumn{1}{c|}{0.367}          & 0.194              & 0.357             \\
			\multirow{-4}{*}{\rotatebox{90}{ETTh2}}  & 720  & 0.217          & \multicolumn{1}{c|}{0.370}          & \textbf{0.173} & \multicolumn{1}{c|}{\textbf{0.342}} & {\ul 0.207} & \multicolumn{1}{c|}{{\ul 0.362}} & 0.320 & \multicolumn{1}{c|}{0.458} & 0.223               & 0.38               & 0.276          & \multicolumn{1}{c|}{0.426}          & 0.234              & 0.388             \\ \hline
			& 96   & {\ul 0.027}    & \multicolumn{1}{c|}{\textbf{0.123}} & 0.036          & \multicolumn{1}{c|}{{\ul 0.146}}    & 0.037       & \multicolumn{1}{c|}{{\ul 0.146}} & 0.075 & \multicolumn{1}{c|}{0.218} & \textbf{0.026}      & \textbf{0.123}     & 0.028          & \multicolumn{1}{c|}{\textbf{0.123}} & 0.028              & 0.129             \\
			& 192  & \textbf{0.039} & \multicolumn{1}{c|}{\textbf{0.151}} & 0.060          & \multicolumn{1}{c|}{0.185}          & 0.053       & \multicolumn{1}{c|}{{\ul 0.180}} & 0.192 & \multicolumn{1}{c|}{0.376} & {\ul 0.04}          & \textbf{0.151}     & 0.045          & \multicolumn{1}{c|}{0.156}          & {\ul 0.04}         & {\ul 0.154}       \\
			& 336  & \textbf{0.051} & \multicolumn{1}{c|}{{\ul 0.175}}    & 0.072          & \multicolumn{1}{c|}{0.204}          & 0.069       & \multicolumn{1}{c|}{0.202}       & 0.146 & \multicolumn{1}{c|}{0.313} & {\ul 0.053}         & \textbf{0.174}     & 0.061          & \multicolumn{1}{c|}{0.182}          & {\ul 0.053}        & 0.177             \\
			\multirow{-4}{*}{\rotatebox{90}{ETTm1}}  & 720  & 0.086          & \multicolumn{1}{c|}{0.226}          & 0.081          & \multicolumn{1}{c|}{0.220}          & 0.089       & \multicolumn{1}{c|}{0.235}       & 0.087 & \multicolumn{1}{c|}{0.230} & {\ul 0.073}         & \textbf{0.206}     & {\ul 0.08}     & \multicolumn{1}{c|}{{\ul 0.21}}     & \textbf{0.072}     & \textbf{0.206}    \\ \hline
			& 96   & \textbf{0.063} & \multicolumn{1}{c|}{{\ul 0.184}}    & 0.079          & \multicolumn{1}{c|}{0.217}          & 0.087       & \multicolumn{1}{c|}{0.223}       & 0.094 & \multicolumn{1}{c|}{0.240} & {\ul 0.065}         & 0.187              & \textbf{0.063} & \multicolumn{1}{c|}{\textbf{0.183}} & 0.071              & 0.203             \\
			& 192  & \textbf{0.087} & \multicolumn{1}{c|}{\textbf{0.220}} & 0.124          & \multicolumn{1}{c|}{0.274}          & 0.135       & \multicolumn{1}{c|}{0.283}       & 0.122 & \multicolumn{1}{c|}{0.276} & 0.093               & 0.231              & {\ul 0.092}    & \multicolumn{1}{c|}{{\ul 0.227}}    & 0.098              & 0.240             \\
			& 336  & \textbf{0.116} & \multicolumn{1}{c|}{\textbf{0.260}} & 0.183          & \multicolumn{1}{c|}{0.337}          & 0.157       & \multicolumn{1}{c|}{0.311}       & 0.179 & \multicolumn{1}{c|}{0.336} & 0.121               & 0.266              & {\ul 0.119}    & \multicolumn{1}{c|}{{\ul 0.261}}    & 0.126              & 0.273             \\
			\multirow{-4}{*}{\rotatebox{90}{ETTm2}}  & 720  & \textbf{0.169} & \multicolumn{1}{c|}{\textbf{0.314}} & 0.188          & \multicolumn{1}{c|}{0.345}          & 0.194       & \multicolumn{1}{c|}{0.351}       & 0.205 & \multicolumn{1}{c|}{0.354} & {\ul 0.172}         & 0.322              & 0.175          & \multicolumn{1}{c|}{{\ul 0.32}}     & 0.180              & 0.332             \\ \hline
			\multicolumn{2}{c|}{Count}      & \multicolumn{2}{c|}{\textbf{16}/{\ul 10}}            & \multicolumn{2}{c|}{\textbf{8}/{\ul 3}}                                                              & \multicolumn{2}{c|}{\textbf{0}/{\ul 4}}                                                        & \multicolumn{2}{c|}{0}                                                             & \multicolumn{2}{c|}{\textbf{8}/{\ul 14}} & \multicolumn{2}{c|}{\textbf{8}/{\ul 8}}              & \multicolumn{2}{c}{\textbf{2}/{\ul 3}} \\ \hline
		\end{tabular}
	}
\end{table*}

\begin{table}[]
    \caption{Comparison of forecast results with and without implicit temporal features.}
    \label{tab:implicit}
    \centering
    \begin{tabular}{c|c|c|c|c}
        \hline
        \multirow{2}{*}{Datasets} & \multicolumn{2}{c|}{Without Pre-train} & \multicolumn{2}{c}{With Pre-train}                                   \\
                                  & MSE                                    & MAE                                & MSE            & MAE            \\ \hline
        ETTh1                     & 0.368                                  & 0.386                              & \textbf{0.359} & \textbf{0.379} \\
        ETTh2                     & 0.272                                  & 0.331                              & \textbf{0.265} & \textbf{0.327} \\
        ETTm1                     & 0.310                                  & 0.351                              & \textbf{0.296} & \textbf{0.338} \\
        ETTm2                     & 0.165                                  & 0.252                              & \textbf{0.160} & \textbf{0.244} \\ \hline
    \end{tabular}
\end{table}

\subsection{Univariate Forecasting}Table~\ref{tab:expuni} showcases univariate long-term time series forecast outcomes for LiPFormer and competing baselines across the entire ETT benchmark datasets. Notably, among the 32 evaluation metrics, LiPFormer ranks within the top-two in 26 metrics and achieves the best performance in 16 of them. This outstanding performance highlights the superiority of LiPFormer over other methods. These findings further demonstrate the robustness of LiPFormer and reconfirm the capability of attention mechanism in time series forecasting, irrespective of multivariate or univariate contexts.

\subsection{More Analysis}
\subsubsection{\bfseries Model Efficiency} We present a thorough comparison of our model's parameters count, training speed, inference time and memory consumption against existing time series forecasting models, using official configurations and identical batch sizes. Tables~\ref{tab:expriments} illustrate the efficiency comparisons under multivariate datasets.

Both LiPFormer and PatchTST significantly outperform the Vanilla Transformer-based models, due to the patching technique. Compared to the state-of-the-art patch-wise PatchTST, LiPFormer still reduces training and inference time by 57\% and 51\% on average, while dramatically decreasing the amounts of computation and model parameters. This margin is especially evident in datasets Traffic and ETTh2, where LiPFormer requires only 2\% MACs compared to PatchTST. The reason for the significant increase in efficiency is twofold. On one hand, we remove computationally intensive components, like LN and FFNs, of the transformer that do not substantially contribute to accuracy. On the other hand, the novel Inter-Patch and Cross-Patch attention mechanisms have not incurred a prohibitive increase in model complexity. Benefitting from its lightweight design, LiPFormer has even better efficiency values than
the linear model TiDE. Although DLinear slightly leads in efficiency with its simple linear structure, the inferior prediction accuracy greatly downgrades its availability.

As in Table~\ref{tab:edge}, we evaluate the compatibility of LiPFormer in resource-constrained environments. To simulate the numerous early-stage or low-budget edge devices already in operation, which typically lack GPUs, we configure a CPU-only edge device with 16GB RAM, 6 cores, and 12 threads. We deploy the trained LiPFormer and Transformer models on this device and conduct prediction inference on the ETTh1 and Weather datasets. Compared to Transformer, LiPFormer exhibit a significant drop in inference time, achieving nearly a tenfold increase in efficiency for an input length of 336 in ETTh1. For the Weather dataset with more channels (21), LiPFormer's efficiency improvement is reduced but still maintains nearly a two-fold enhancement. Notably, Transformer exceeds the memory limit for an input length of 720 on both datasets, whereas LiPFormer exhibits its lightweight superiority.
\begin{table}[]
    \centering
    \caption{Comparison of inference time (seconds per inference) in a CPU-only device varying input sequence lengths.}
    \label{tab:edge}
    \addtolength{\tabcolsep}{-3pt}
    \resizebox{0.9\linewidth}{!}{\begin{tabular}{c|cccc|cccc}
		\hline
	\multirow{2}{*}{Input Lengths}			 & \multicolumn{4}{c|}{ETTh1}  & \multicolumn{4}{c}{Weather} \\
				 & 96    & 192  & 336  & 720  & 96    & 192   & 336  & 720  \\ \hline
	Transformer  & 1.47  & 2.86 & 5.82 & -    & 1.84  & 3.88  & 6.08 & -    \\
	LipFormer    & 0.55  & 0.60 & 0.62 & 0.74 & 0.87  & 1.35  & 3.77 & 4.57 \\ \hline
	\end{tabular}}
\end{table}
\subsubsection{\bfseries Impact of Patch Size} To verify the impact of patch length $pl$, we perform experiments with four patch lengths. Results are shown on Table~\ref{tab:ab-pl}. Fixed patch length does not lead to its performance loss on different datasets, which proves that the mixing operation we adopted effectively improves the generalization performance. Specifically, We recommend using $pl=48$ as a more suitable choice for most datasets.
\begin{table*}
    \caption{The impact of patch size. $pl=\{6,12,24,48\}$ for all datasets. }
    \label{tab:ab-pl}
    \centering
    \addtolength{\tabcolsep}{-2pt}
    \scriptsize\resizebox{0.95\textwidth}{!}{
        \begin{tabular}{c|c|cccc|cccc|cccc|cccc}
            \hline
            Datasets                 & \multirow{2}{*}{Metric} & \multicolumn{4}{c|}{ETTh1} & \multicolumn{4}{c|}{ETTh2} & \multicolumn{4}{c|}{ETTm1} & \multicolumn{4}{c}{ETTm2}                                                                                                                                                                                                             \\
            Forecasting Length       &                         & 96                        & 192                        & 336                        & 720                        & 96             & 192            & 336            & 720            & 96             & 192            & 336            & 720            & 96             & 192            & 336            & 720            \\
            \hline
            \multirow{2}{*}{pl = 6}  & MSE                     & 0.372                     & 0.405                      & \textbf{0.427}             & \textbf{0.439}             & 0.276          & \textbf{0.333} & 0.365          & 0.402          & 0.482          & \textbf{0.440} & 0.422          & 0.417          & 0.575          & 0.375          & 0.358          & \textbf{0.369} \\
                                     & MAE                     & 0.397                     & 0.413                      & 0.430                      & 0.452                      & 0.337          & 0.374          & 0.400          & 0.435          & 0.440          & 0.426          & 0.412          & 0.411          & \textbf{0.390} & 0.382          & 0.373          & \textbf{0.378} \\ \hline
            \multirow{2}{*}{pl=12}   & MSE                     & 0.375                     & 0.417                      & 0.431                      & 0.468                      & 0.268          & 0.335          & 0.364          & \textbf{0.390} & 0.480          & 0.442          & \textbf{0.420} & \textbf{0.414} & 0.402          & 0.370          & \textbf{0.356} & 0.376          \\
                                     & MAE                     & 0.395                     & 0.415                      & 0.431                      & 0.463                      & 0.333          & 0.376          & 0.398          & 0.426          & 0.439          & \textbf{0.424} & \textbf{0.411} & \textbf{0.411} & 0.392          & 0.379          & \textbf{0.372} & 0.381          \\ \hline
            \multirow{2}{*}{pl = 24} & MSE                     & 0.369                     & 0.411                      & 0.423                      & 0.450                      & 0.274          & 0.336          & 0.366          & 0.391          & \textbf{0.479} & 0.446          & 0.426          & 0.507          & 0.403          & \textbf{0.370} & 0.358          & 0.370          \\
                                     & MAE                     & 0.387                     & 0.407                      & 0.437                      & \textbf{0.448}             & 0.335          & \textbf{0.373} & 0.399          & 0.430          & \textbf{0.436} & 0.426          & 0.412          & 0.456          & 0.393          & \textbf{0.379} & 0.374          & 0.378          \\ \hline
            \multirow{2}{*}{pl=48}   & MSE                     & \textbf{0.359}            & \textbf{0.404}             & 0.444                      & 0.450                      & \textbf{0.265} & 0.335          & \textbf{0.364} & 0.392          & 0.483          & 0.450          & 0.425          & 0.526          & \textbf{0.400} & 0.372          & 0.357          & 0.372          \\
                                     & MAE                     & \textbf{0.379}            & \textbf{0.405}             & \textbf{0.425}             & 0.453                      & \textbf{0.327} & 0.374          & \textbf{0.395} & \textbf{0.425} & 0.439          & 0.429          & 0.412          & 0.473          & 0.391          & 0.380          & 0.374          & 0.380          \\
            \hline
        \end{tabular}}
\end{table*}

\subsubsection{\bfseries Impact of Input Length}Since longer input sequences indicate that more historical information is available, models with strong ability to capture long-term time dependencies should perform better when the input length increases. As shown in Table~\ref{tab:inputl}, a comprehensive evaluation on the MSE metric is conducted using the ETT and Weather datasets with varying input lengths (96, 192, 336, 720). Overall, the performance of LiPFormer improves as input length increases. LipFormer significantly outperforms SOTA benchmarks under most input lengths (15/20), indicating that our model can effectively extract useful information from histories and capture long-term dependencies.

\begin{table}[]
	\centering
	\caption{Impact of input sequence length.}
	\label{tab:inputl}
	\addtolength{\tabcolsep}{-5pt}
	\scriptsize\resizebox{\linewidth}{!}{\begin{tabular}{cc|c|c|c|c|c|c|c}
			\hline
			Datasets                                 & Input length & \textbf{LipFormer} & PatchTST       & DLinear & TiDE  & iTransformer & FGNN  & TimeMixer      \\ \hline
			\multirow{4}{*}{\rotatebox{90}{ETTh1}}   & 96           & \textbf{0.373}     & 0.383          & 0.396   & 0.432 & 0.394        & 0.503 & 0.398          \\
			& 192          & \textbf{0.368}     & 0.380          & 0.386   & 0.424 & 0.396        & 0.530 & 0.454          \\
			& 336          & 0.383              & \textbf{0.382} & 0.375   & 0.410 & 0.397        & 0.553 & 0.393          \\
			& 720          & \textbf{0.359}     & 0.375          & 0.375   & 0.375 & 0.392        & 0.647 & 0.385          \\ \hline
			\multirow{4}{*}{\rotatebox{90}{ETTh2}}   & 96           & \textbf{0.286}     & 0.317          & 0.341   & 0.318 & 0.300        & 0.416 & 0.286          \\
			& 192          & \textbf{0.287}     & 0.311          & 0.323   & 0.311 & 0.302        & 0.409 & 0.294          \\
			& 336          & \textbf{0.277}     & 0.308          & 0.307   & 0.297 & 0.307        & 0.390 & 0.287          \\
			& 720          & \textbf{0.265}     & 0.274          & 0.289   & 0.270 & 0.303        & 0.479 & 0.296          \\ \hline
			\multirow{4}{*}{\rotatebox{90}{ETTm1}}   & 96           & 0.326              & 0.335          & 0.345   & 0.365 & 0.341        & 0.400 & \textbf{0.324} \\
			& 192          & 0.306              & 0.310          & 0.306   & 0.319 & 0.303        & 0.368 & \textbf{0.302} \\
			& 336          & \textbf{0.285}     & 0.293          & 0.300   & 0.310 & 0.304        & 0.378 & 0.298          \\
			& 720          & 0.296              & \textbf{0.290} & 0.299   & 0.306 & 0.318        & 0.403 & 0.305          \\ \hline
			\multirow{4}{*}{\rotatebox{90}{ETTm2}}   & 96           & \textbf{0.175}     & 0.182          & 0.193   & 0.188 & 0.183        & 0.230 & 0.176          \\
			& 192          & \textbf{0.167}     & 0.175          & 0.179   & 0.176 & 0.185        & 0.231 & 0.171          \\
			& 336          & \textbf{0.161}     & 0.172          & 0.169   & 0.170 & 0.173        & 0.223 & 0.179          \\
			& 720          & \textbf{0.160}     & 0.165          & 0.161   & 0.167 & 0.180        & 0.225 & 0.181          \\ \hline
			\multirow{4}{*}{\rotatebox{90}{Weather}} & 96           & 0.179              & \textbf{0.177} & 0.196   & 0.177 & 0.184        & 0.170 & 0.184          \\
			& 192          & 0.162              & \textbf{0.159} & 0.184   & 0.187 & 0.168        & 0.182 & 0.168          \\
			& 336          & 0.154              & \textbf{0.15}  & 0.174   & 0.172 & 0.158        & 0.173 & 0.158          \\
			& 720          & \textbf{0.146}     & 0.152          & 0.176   & 0.166 & 0.159        & 0.166 & 0.151          \\ \hline
	\end{tabular}}
\end{table}
\subsubsection{\bfseries Ablation of Lightweight Architecture} \label{ssc:LA} Time series data does not have a large number of data entries and the corresponding inductive bias as in the NLP domain, the Normalization approach used by the Transformer class of models and their deeper layers may capture greater noise in time series prediction, In order to validate the effectiveness of our deletion of the layer norm versus the feed forward layer, the we constructed four model model variants without removing these parts:
\begin{itemize}
    \item {\bfseries LiPFormer + LN:} LN after adding the Attention mechanism to the original model
    \item  {\bfseries LiPFormer + FFNs:} FeedForward network added to the Attention mechanism and then exported
    \item {\bfseries LiPFormer + LN + FFNs:} Add all the above components to the LiPFormer model.
\end{itemize}

The comparison experiments are shown in Table~\ref{tab:diffmcl}. The use of either FFNs or LN caused a decrease in the performance of LiPFormer, where the average MSE and MAE of LiPFormer increased by 15 \% and 9\%, respectively, after the addition of FFNs, while the addition of LN also had an impact on the performance of LiPFormer, with an increase in the MSE by 3.5\% and in the MAE by 4\%. In particular, after adding both FFNs and LN, LiPFormer's MSE improves by 24\% on average and MAE improves by 13\% on average, and this severe performance degradation in the ETTh1 dataset results in a 45\% improvement in MSE and a 23\% improvement in MAE. This experimental result is consistent with our assumption that FFNs and LN are not effective for time series modeling. In fact, removing these components tends to contribute to the accuracy improvement of the time series prediction.
\begin{table}
    \centering
    \caption{Ablation study of architecture lightweight, including Feed Forward Networks and Layer Normalization.}
    \label{tab:diffmcl}
    \addtolength{\tabcolsep}{-2pt}
    \scriptsize\resizebox{\linewidth}{!}{
        \begin{tabular}{c|c|cccc|cccc}
            \hline
            \multirow{2}{*}{Datasets}  & \multirow{2}{*}{Metric} & \multicolumn{4}{c|}{ETTh1} & \multicolumn{4}{c}{ETTm2}                                                 \\
                                       &                         & 96                        & 192                        & 336   & 720   & 96    & 192   & 336   & 720   \\
            \hline
            LiPFormer                  & MSE                     & 0.389                     & 0.439                      & 0.514 & 0.523 & 0.184 & 0.217 & 0.282 & 0.521 \\
            +FFNs                      & MAE                     & 0.408                     & 0.441                      & 0.478 & 0.497 & 0.271 & 0.285 & 0.330 & 0.466 \\ \hline
            LiPFormer                  & MSE                     & 0.392                     & 0.437                      & 0.448 & 0.508 & 0.158 & 0.218 & 0.266 & 0.350 \\
            +LN                        & MAE                     & 0.412                     & 0.435                      & 0.453 & 0.498 & 0.243 & 0.286 & 0.317 & 0.376 \\ \hline
            LiPFormer                  & MSE                     & 0.754                     & 0.505                      & 0.614 & 0.498 & 0.164 & 0.226 & 0.284 & 0.360 \\
            +FFNs+LN                   & MAE                     & 0.557                     & 0.465                      & 0.524 & 0.497 & 0.252 & 0.292 & 0.332 & 0.381 \\ \hline
            \multirow{2}{*}{LiPFormer} & MSE                     & 0.359                     & 0.404                      & 0.444 & 0.450 & 0.160 & 0.217 & 0.273 & 0.348 \\
                                       & MAE                     & 0.379                     & 0.405                      & 0.425 & 0.453 & 0.244 & 0.285 & 0.322 & 0.372 \\
            \hline
        \end{tabular}}

\end{table}
\begin{table*}[]
    \centering
    \caption{Ablation study of patch-wise attentions, including Cross-patch attention and Inter-patch attention.}
    \label{tab:diff-attn}
    \addtolength{\tabcolsep}{-2pt}
    \scriptsize\resizebox{0.95\linewidth}{!}{
        \begin{tabular}{cc|cccc|cccc|cccc|cccc}
            \hline
            \multirow{2}{*}{Datasets}    & \multirow{2}{*}{Metric} & \multicolumn{4}{c|}{ETTh1} & \multicolumn{4}{c|}{ETTh2} & \multicolumn{4}{c|}{ETTm1} & \multicolumn{4}{c}{ETTm2}                                                                                                          \\
                                         &                         & 96                        & 192                        & 336                        & 720                        & 96             & 192   & 336   & 720   & 96    & 192   & 336   & 720   & 96    & 192   & 336   & 720   \\
            \hline
            Without                      & MSE                     & 0.366                     & \textbf{0.401}             & \textbf{0.441 }            & 0.461                      & 0.270          & 0.332 & 0.370 & 0.394 & 0.327 & 0.348 & 0.386 & 0.434 & 0.167 & 0.225 & 0.280 & 0.364 \\
            Cross-Patch attn.            & MAE                     & 0.380                     & 0.424                      & 0.450                      & 0.461                      & 0.329          & 0.371 & 0.396 & 0.427 & 0.365 & 0.378 & 0.398 & 0.425 & 0.255 & 0.295 & 0.333 & 0.386 \\ \hline
            Without                      & MSE                     & 0.364                     & 0.413                      & 0.458                      & 0.472                      & 0.271          & 0.335 & 0.383 & 0.586 & 0.309 & 0.342 & 0.370 & 0.430 & 0.164 & 0.220 & 0.275 & 0.353 \\
            Inter-Patch attn.            & MAE                     & 0.380                     & 0.408                      & 0.434                      & 0.455                      & 0.329          & 0.371 & 0.405 & 0.477 & 0.350 & 0.367 & 0.384 & 0.417 & 0.251 & 0.289 & 0.324 & 0.382 \\ \hline
            \multirow{2}{*}{Neither} & MSE                     & 0.379                     & 0.417                      & 0.456                      & 0.466                      & 0.273          & 0.379 & 0.397 & 0.422 & 0.314 & 0.340 & 0.377 & 0.423 & 0.168 & 0.224 & 0.276 & 0.356 \\
                                         & MAE                     & 0.392                     & 0.412                      & 0.432                      & 0.453                      & 0.332          & 0.388 & 0.427 & 0.441 & 0.356 & 0.371 & 0.393 & 0.419 & 0.255 & 0.293 & 0.330 & 0.379 \\ \hline
            \multirow{2}{*}{LiPFormer}   & MSE                     & \textbf{0.359 }           & 0.404                      & 0.444                      & \textbf{0.450}             & \textbf{0.265} & \textbf{0.330} & \textbf{0.364} & \textbf{0.392} & \textbf{0.296} & \textbf{0.336} & \textbf{0.365} & \textbf{0.408} & \textbf{0.160} & \textbf{0.217 }& \textbf{0.273} & \textbf{0.348} \\
                                         & MAE                     & \textbf{0.379}            & \textbf{0.405}             & \textbf{0.425}             & \textbf{0.453}             & \textbf{0.327} & \textbf{0.369} & \textbf{0.395} & \textbf{0.425 }& \textbf{0.338} & \textbf{0.360} & \textbf{0.379} & \textbf{0.413} & \textbf{0.244} & \textbf{0.285} & \textbf{0.322} & \textbf{0.372} \\ \hline
        \end{tabular}}
\end{table*}
\begin{figure}
    \centering
    \includegraphics[width=0.23\textwidth]{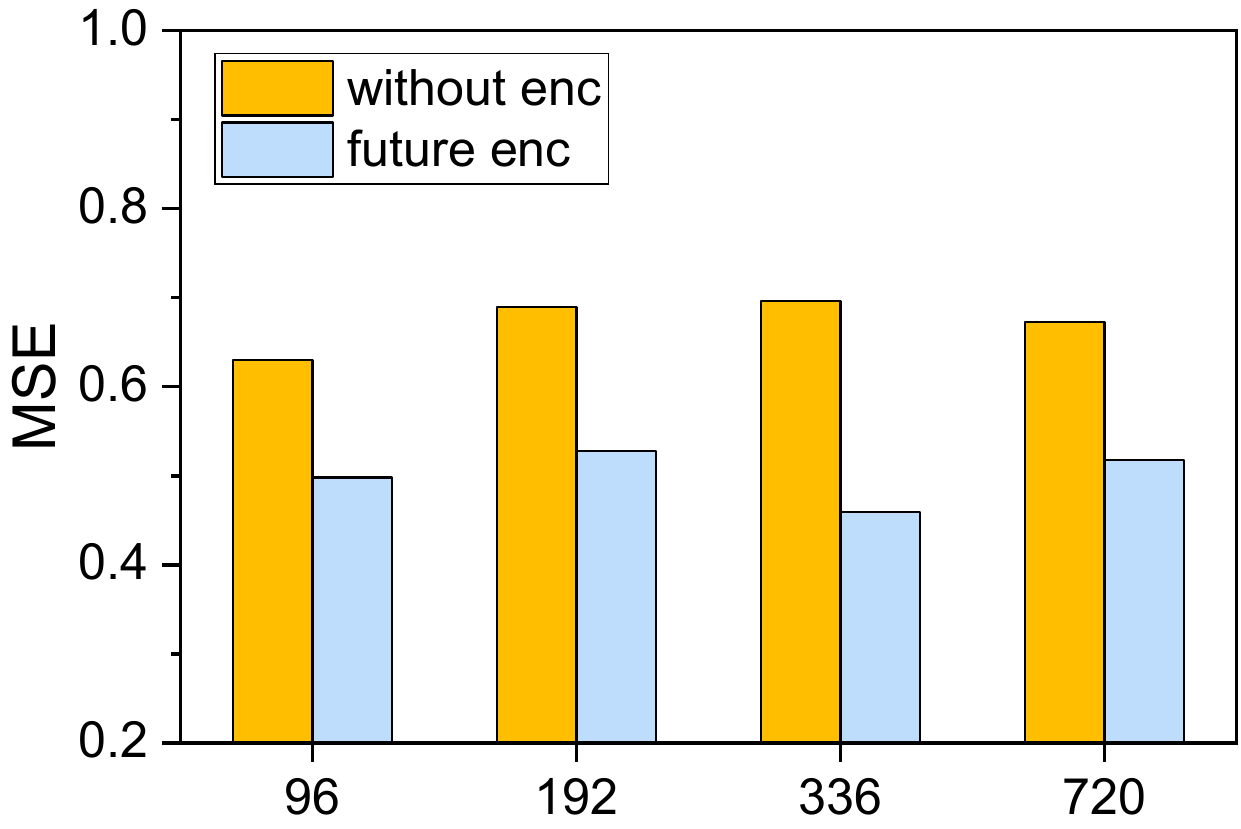}
    \includegraphics[width=0.23\textwidth]{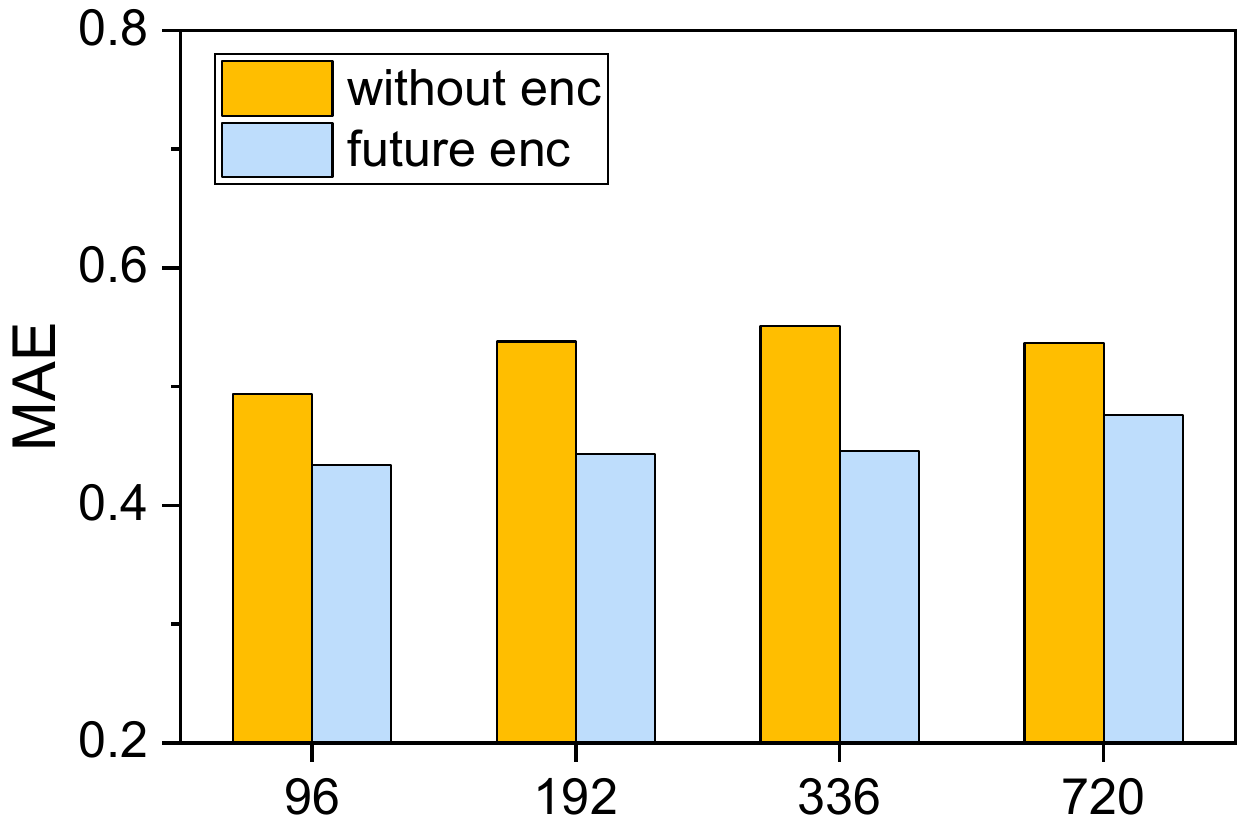}
    \caption{The impact of incorporating or excluding the future Covariate Encoder (enc) on the MSE and MAE in the Electricity Price dataset.}
    \label{fig:abenc}
\end{figure}

\begin{figure}
	\centering
	\includegraphics[width=0.4\linewidth]{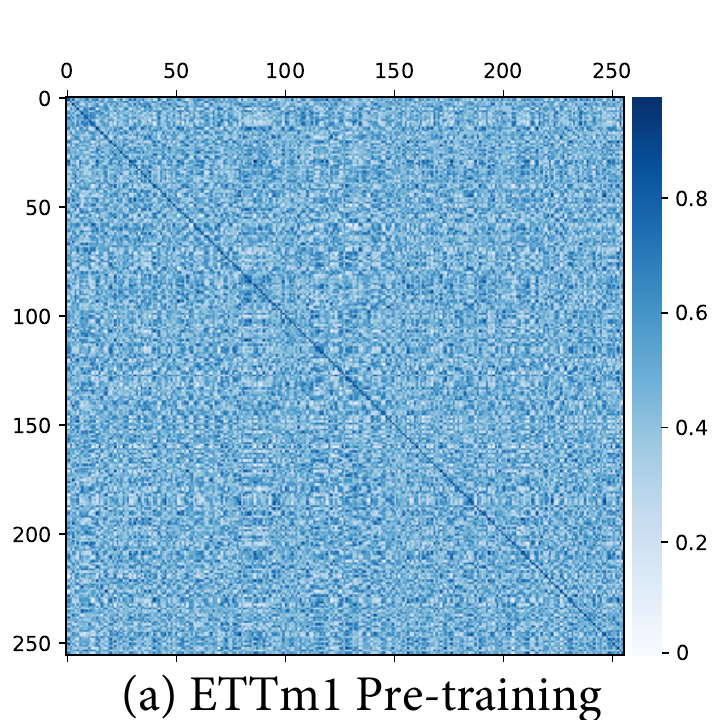}
	\includegraphics[width=0.4\linewidth]{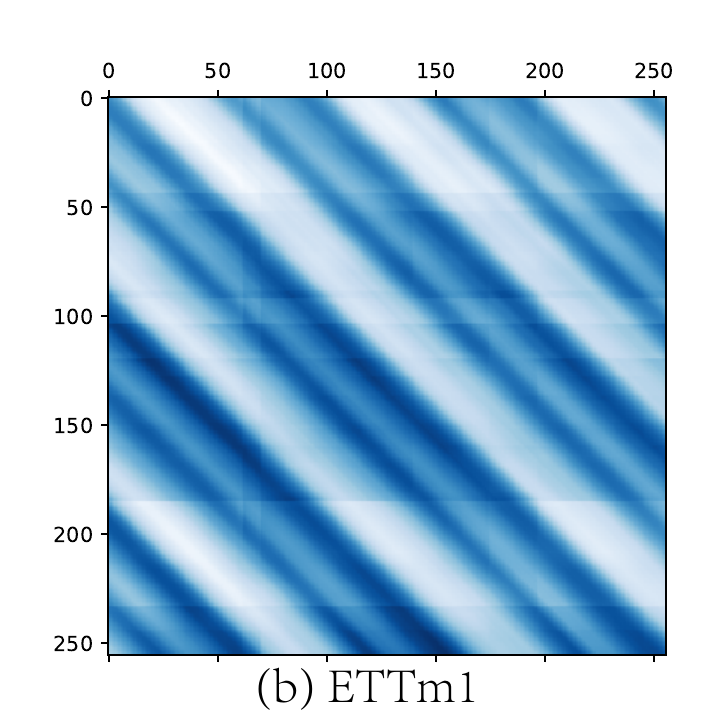}
	\includegraphics[width=0.4\linewidth]{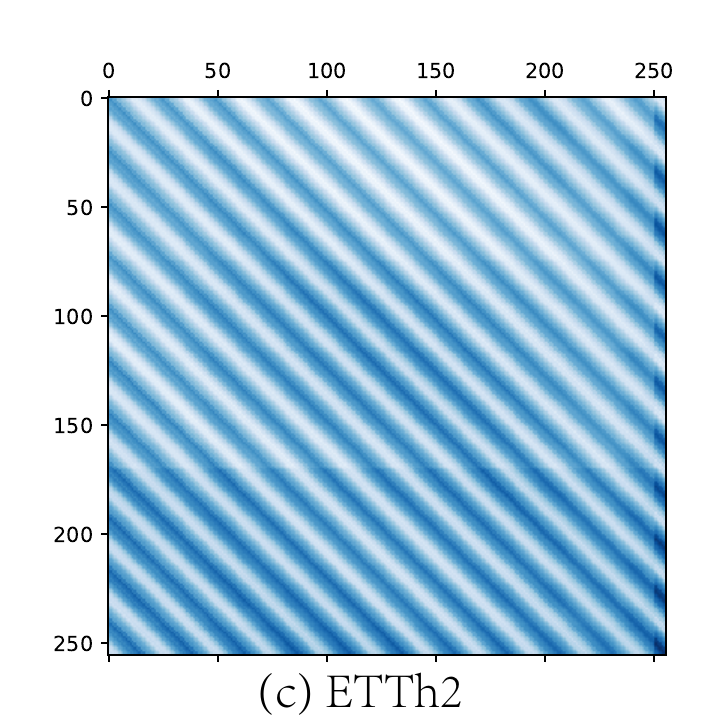}
	\includegraphics[width=0.4\linewidth]{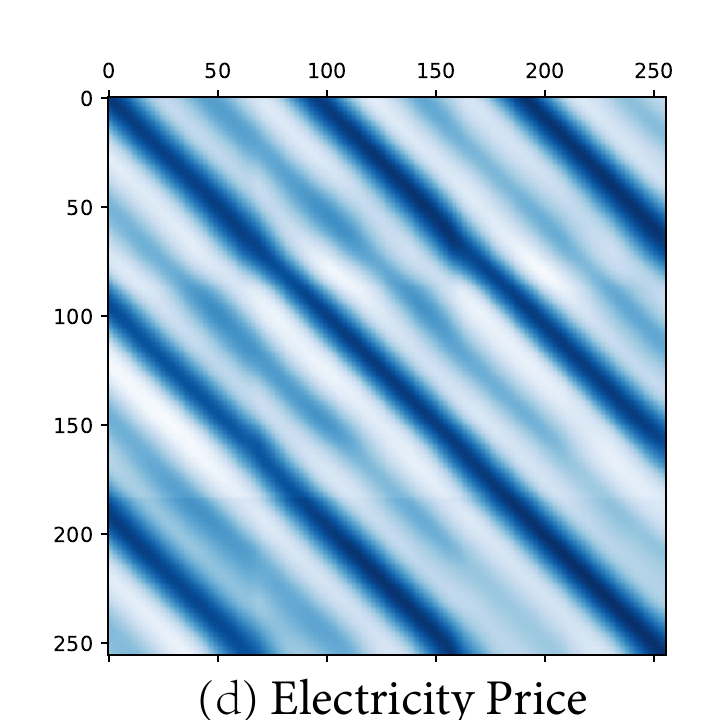} 
	\caption{Visualization of the logits matrices for Weakly Supervised Architecture: (a) Logits matrix pre-trained on ETTm1; (b), (c), (d) Logits matrices on the validation sets of ETTm1, ETTh2 and Electricity-Price, where $b=256$.}
	\label{fig:vis}
\end{figure}

\subsubsection{\bfseries Ablation of Patch-wise Attention}To verify the effectiveness of Cross- and Inter-Patch attention mechanisms, we constructed three model variants for comparison:
\begin{itemize}
    \item {\bfseries Without Cross-Patch attn.:} We remove the Cross-Patch attention and use a linear layer instead.
    \item {\bfseries Without Inter-Patch attn.:} We replace the Inter-Patch attention with a linear layer.
    \item {\bfseries Neither:} Only traditional patching technique is used.
\end{itemize}

Table~\ref{tab:diff-attn} reports the impact of altering the patching mechanisms. Compared to using only the classical patching method, employing Cross-Patch alone outperforms using Inter-Patch alone. The former exceeds Neither in 72\% of the metrics, while the latter shows only comparable results. This may be attributed to Cross-Patch's ability to mitigate fixed patch size limitation, which possibly makes Inter-Patch less effective on ETTm than on ETTh. It should be noted that Cross-Patch and Inter-Patch are designed to be complementary in their effects, serving to perceive global trends and local correlations, respectively. While using either mechanism alone results in only limited improvement, their combined use consistently demonstrates robust accuracy enhancements across all datasets, with MSE and MAE decreasing by 5\% and 3\%, respectively. This validates the essentiality of combining both mechanisms and their collective effectiveness.

\begin{table}[]
    \centering
    \caption{Covariate encoder to other models.}
    \label{tab:fcom}
    \scriptsize\resizebox{\linewidth}{!}{\begin{tabular}{cc|cc|cc}
            \hline
            \multirow{2}{*}{Models}      &     & \multicolumn{2}{c|}{Covariate Encoder} & \multicolumn{2}{c}{Without Covariate Encoder}                 \\
                                         &     & MSE                                   & MAE                                            & MSE   & MAE   \\
            \hline
            \multirow{2}{*}{Informer}    & 96  & \textbf{0.699}                        & \textbf{0.525}                                 & 0.736 & 0.569 \\
                                         & 192 & \textbf{0.692}                        & \textbf{0.569}                                 & 0.706 & 0.576 \\ \hline
            \multirow{2}{*}{Transformer} & 96  & \textbf{0.691}                        & \textbf{0.530}                                 & 0.723 & 0.571 \\
                                         & 192 & \textbf{0.691}                        & \textbf{0.533}                                 & 0.704 & 0.549 \\ \hline
            \multirow{2}{*}{Autoformer}  & 96  & \textbf{0.699}                        & \textbf{0.583}                                 & 0.739 & 0.592 \\
                                         & 192 & \textbf{0.685}                        & \textbf{0.536}                                 & 0.744 & 0.580 \\ \hline
        \end{tabular}}

\end{table}
\subsubsection{\bfseries Ablation of Covariate Encoder}We verify the effect of the future covariate encoder in a simple way: as shown in the Figure~\ref{fig:abenc}, by removing the covariate encoder, in the dataset with future covariates, the MSE of the LiPFormer decreases by 34\%, and the MAE decreases by 17\%, but it still outperforms the SOTA model in a number of cases, which is side by side a proof of the validity of our underlying predictor.

We further integrate the Covariate Encoder seamlessly into various models, including Transformer, Informer, and Autoformer. To substantiate the efficacy of this encoder, we conduct experiments on the Electricity-Price dataset. The experimental results reported in Table~\ref{tab:fcom} reveal that all the transformer-based models incorporating the Covariate Encoder outperform their original versions, achieving an average reduction of 4\% in MSE and 5\% in MAE. These improvements confirm both the validity of our proposed inductive bias about weak labels and the effectiveness of the Dual Encoder architecture.

\subsubsection{\bfseries Visiualization}Figure~\ref{fig:vis} visualizes the logits matrices across various datasets, revealing that our weakly supervised learning approach aligns with the latent vectors between predictions (X-axis) and future covariates (Y-axis). The $b$ diagonal values in Figure~\ref{fig:vis}(a) highlight how contrastive learning optimizes similarity for true values. Given that validation sets are unshuffled, Figures~\ref{fig:vis}(b) and (c) display periodic correlations in the logits matrices corresponding to the actual periods (ETTm1=96, ETTh2=24). For datasets featuring explicit covariates, as shown in Figure~\ref{fig:vis}(d), we observe clear periodicity alongside irregularities at the edges of ``stripes'', including less pronounced ``blurred stripes'' within these periods. These patterns support the role of explicit weak labels in guiding predictions, consistent with our inductive bias that future covariates correlate with time series models. Due to space constraints, similar results from other datasets are not included.

\section{CONCLUSION} \label{se:cc}

This paper proposes a Lightweight Patch-wise Transformer with weak label enriching (LiPFormer) for time series forecasting. To simplify the Transformer backbone, it integrates a novel Cross-Patch mechanism into existing patching attention and devises a linear transformation-based attention to eliminate Positional Encoding and two heavy components, Layer Normalization and Feed Forward Networks. A weak label enriching architecture is presented to leverage valuable context information for modeling multimodel future covariates via a dual encoder contrastive learning framework. These innovative mechanisms collectively make LiPFormer a significantly lightweight model, achieving outstanding prediction performance. Deployment on an CPU-only edge device and transplant trials of the weak label enriching module further demonstrate the scalability and versatility of LiPFormer.
\section*{Acknowledgments}
This work was supported by National Natural Science Foundation of China (No. 62272369, 62372352), Natural Science Basic Research Program of Shaanxi (No. 2023-JC-YB-558, 2024JC-YBMS-473), Scientific Research Program Funded by Shaanxi Provincial Education Department (No. 23JS028), Scientific and Technological Program of Xi'an (No. 24GXFW0016).

\clearpage 

\bibliographystyle{IEEEtran}
\balance
\bibliography{IEEEabrv,LiPFormer}

\end{document}